\documentclass[lettersize,journal]{IEEEtran}
\usepackage{amsmath,amsfonts}
\usepackage{algorithmic}
\usepackage{algorithm}
\usepackage{array}
\usepackage[caption=false,font=normalsize,labelfont=sf,textfont=sf]{subfig}
\usepackage{textcomp}
\usepackage{stfloats}
\usepackage{url}
\usepackage{verbatim}
\usepackage{graphicx}
\usepackage{cite}
\usepackage{booktabs}
\usepackage[table]{xcolor}
\usepackage{multirow}
\usepackage{array}
\usepackage{colortbl}
\usepackage{afterpage}
\usepackage{float}
\usepackage[backref]{hyperref} 
\hypersetup{hidelinks}
\usepackage{colortbl}  
\usepackage{xcolor}
\usepackage{amsmath}
\usepackage{amssymb}
\usepackage{color}
\usepackage{indentfirst}
\usepackage{threeparttable}
\usepackage{pifont}
\usepackage{makecell}
\usepackage{arydshln}

\begin{document}

\title{DHQA-4D: Perceptual Quality Assessment of Dynamic 4D Digital Human}

\author{Yunhao Li, Sijing Wu, Yucheng Zhu, Huiyu Duan, Zicheng Zhang,\\ Guangtao Zhai,~\IEEEmembership{Fellow,~IEEE}

\thanks{Yunhao Li, Sijing Wu, Huiyu Duan, Zicheng Zhang, Guangtao Zhai are with the Institute of Image Communication and Network Engineering, Shanghai Jiao Tong University, Shanghai, China (e-mail: \{lyhsjtu, wusijing, huiyuduan, zzc1998, zhaiguangtao\}@sjtu.edu.cn).}

\thanks{Yucheng Zhu is USC-SJTU Institute of Cultural and Creative Industry, Shanghai Jiao Tong University, Shanghai, China (e-mail: zyc420@sjtu.edu.cn).}
}

\maketitle
\begin{abstract}

With the rapid development of 3D scanning and reconstruction technologies, dynamic digital human avatars based on 4D meshes have become increasingly popular. A high-precision dynamic digital human avatar can be applied to various fields such as game production, animation generation, and remote immersive communication. However, these 4D human avatar meshes are prone to being degraded by various types of noise during the processes of collection, compression, and transmission, thereby affecting the viewing experience of users. In light of this fact, quality assessment of dynamic \underline{4D} digital humans becomes increasingly important. In this paper, we first propose a large-scale dynamic \underline{d}igital \underline{h}uman \underline{q}uality \underline{a}ssessment dataset, DHQA-4D, which contains 32 high-quality real-scanned 4D human mesh sequences, 1920 distorted textured 4D human meshes degraded by 11 textured distortions, as well as their corresponding textured and non-textured mean opinion scores (MOSs). Equipped with DHQA-4D dataset, we analyze the influence of different types of distortion on human perception for textured dynamic 4D meshes and non-textured dynamic 4D meshes. Additionally, we propose DynaMesh-Rater, a novel large multimodal model (LMM) based approach that is able to assess both textured 4D meshes and non-textured 4D meshes. Concretely, DynaMesh-Rater elaborately extracts multi-dimensional features, including visual features from a projected 2D video, motion features from cropped video clips, and geometry features from the 4D human mesh to provide comprehensive quality-related information. Then we utilize a LMM model to integrate the multi-dimensional features and conduct a LoRA-based instruction tuning technique to teach the LMM model to predict the quality scores. Extensive experimental results on the DHQA-4D dataset demonstrate the superiority of our DynaMesh-Rater method over previous quality assessment methods.

\end{abstract}


\section{Introduction}

In recent years, with the rapid development of 3D scanning and reconstruction technologies, high-precision static and dynamic 3D meshes are becoming increasingly popular. Among the high-precision dynamic 3D meshes, \textit{a.k.a} 4D meshes, digital human avatars are of great significance due to their extensive application scenarios in the gaming field, virtual reality (VR), augmented reality (AR), and remote communication. A high-quality dynamic 3D human avatar represented by a mesh sequence can be placed into a VR headset, allowing people to remotely engage in various social activities. However, during the processes of collection, compression, and transmission, the mesh sequence of the 3D human avatar might encounter various distortion problems, which significantly affect the viewing experience of users. For instance, real-time transmission may cause delays or latency, and 3D model compression will introduce quality degradations on shapes and textures. Considering these problems, subjective and objective quality assessment of dynamic 3D human meshes is becoming more and more significant.

Quality assessment of general 3D meshes or 3D digital human meshes has been studied for a long time. However, current works are mainly focused on static meshes and existing dynamic mesh datasets contain limited reference meshes and distortion types. For static meshes, CMDM \cite{nehme2020visual} is the first static mesh quality assessment data set with only 80 distorted meshes, then TMQA \cite{nehme2023textured} constructs the current largest mesh assessment dataset with 3000 distorted samples from 55 source mesh models. As for dynamic meshes, only a few studies have been conducted due to the lack of high-quality 4D meshes before. Yang \emph{et al.} \cite{yang2024tdmd} have proposed a general 4D mesh quality assessment database with 303 samples from 8 reference sequences. Zhang \emph{et al.} \cite{zhang2023ddh} have proposed the first dynamic digital human quality assessment database containing 800 distorted digital humans from only 2 reference sequences. To conduct a comprehensive subjective and objective study for dynamic 4D textured and non-textured digital humans, a large-scale quality assessment dataset is urgently needed. The difficulty in creating large-scale dynamic digital human assessment datasets is that high-precision 4D mesh scan data is very difficult to obtain. However, with the release of more and more open-source high-precision dynamic human mesh datasets, this difficulty is gradually fading.


In light of these facts, we first propose a comprehensive quality assessment dataset \textbf{DHQA-4D} to facilitate dynamic 4D \underline{d}igital \underline{h}uman \underline{q}uality \underline{a}ssessment. The DHQA-4D dataset consists of 32 high-quality real-scanned dynamic digital humans with complex clothing represented by 4D textured meshes. The collected raw dynamic digital human meshes are then degraded by 11 types of distortions including geometry gaussian noise (GN), color noise (CN), texture downsampling (TD), mesh simplification (MS), texture map compression (TMC), position compression (PC), UV map compression (UMC), position and UV Map compression (PUC), geometry and texture compression (GTC), mixed compression (MC) and temporal discontinuity (DC) to generate 1920 distorted meshes. To comprehensively evaluate the quality of 4D meshes, we conduct a subjective evaluation from texture and shape perspectives, respectively. Specifically, we obtain texture-based mean opinion scores (MOSs) from all 1920 distorted 4D meshes with texture, while obtain 832 shape-based MOSs from 832 distorted 4D meshes without textures which are degraded by GN, MS, PC, MC, and DC distortions, respectively, since other distortions have no obvious degradation on shapes. Compared to the previous dynamic mesh datasets, our DHQA-4D dataset conducts a more comprehensive subjective study of the perceptual quality on 4D human avatars under various distortions, which carefully analyzes the influences of single and mixed distortions on human perception from both texture and shape perspectives for the first time. We believe that DHQA-4D can provide valuable insights for optimizing downstream applications such as mesh compression and mesh denoising for dynamic digital humans.

\begin{table*}
\setlength{\belowcaptionskip}{-0.02cm}
  \centering
\renewcommand\arraystretch{1}
\caption{An overview of current 3D/4D mesh quality assessment dataset.}
   \resizebox{0.8\linewidth}{!}{\begin{tabular}{lcccccccc}
    \toprule[1pt]
     \bf Model & \bf Scale &\bf Texture &\bf Content & \bf Distortions & \bf Type & \bf Avg. Vertices/Faces \\
    \midrule

    3DMAQD (SPIC 2015) \cite{torkhani2015perceptual} & 276 & No & General &  4 & dynamic & 40k/80k \\
    CMDM (TVCG 2021) \cite{nehme2020visual} &  80 & Yes & General  & 4  & Static & --/-- \\
    TMQA (TOG 2022) \cite{nehme2023textured} & 3000 & Yes & General & 5 & Static &  150k/300k \\

    SJTU-TMQA (ICASSP 2024) \cite{cui2024sjtu} & 945 & Yes &  General  & 8 & Static &  2.8k/5.6k    \\

    SJTU-H3D (Arxiv 2023) \cite{zhang2023advancing} &  1120 &  Yes & Digital Human & 7 & Static & 20k/40k \\

    DDH-QA (ICME 2023) \cite{zhang2023ddh} & 800  & Yes  & Synthetic Human  & 8  & dynamic  & 1.9k/3.8k \\

    TDMD (TVCG 2024) \cite{yang2024tdmd} & 303 & Yes & General  & 6 & dynamic &  20k/40k \\

    \textbf{DHQA-4D (Ours)} &  1920 & Yes $\&$ No &  Real Scanned Human &  11 &  dynamic  &  40k/80k  \\

    \bottomrule[1pt]
  \end{tabular}}
  \label{4ddataset}
\end{table*}

Equipped with our \textbf{DHQA-4D} dataset, we benchmark the current full reference (FR) and no-reference objective metrics for evaluating dynamic 4D human meshes. Then, focusing on the no-reference (NR) 4D mesh assessment task, utilizing the powerful visual understanding ability of a large multimodal model (LMM), we propose a LMM-based quality assessment method named \textbf{DynaMesh-Rater}. DynaMesh-Rater consists of a visual encoder to extract visual
features from sparse projected video frame, a motion encoder to extract motion features from uniformly cropped video clips, a geometry encoder to extract shape-aware features from the surfaces of dynamic human meshes, a large language model (LLM) to unify the multidimensional features into a compact language feature space, and a quality regressor to output the contiguous quality score. Through utilizing the multidimensional features and a low-rank adaptation (LoRA) technique for instruction tuning. We optimize LLM to learn human preference characteristics and assess the quality of dynamic 4D digital humans. Extensive experimental results demonstrate the effectiveness of our DynaMesh-Rater framework. 

In summary, the overall contributions of our paper are:

\begin{itemize}

\item We establish a large-scale comprehensive perceptual quality assessment dataset \textbf{DHQA-4D} for dynamic 4D digital human meshes with quality scores of both textured dynamic digital human and non-textured dynamic digital human. DHQA-4D consists of 32 high-quality dynamic digital human sequences represented by textured mesh sequences, 1920 distorted textured digital human meshes with MOS annotations generated from 11 types of distortions, and 832 distorted non-textured digital human meshes with MOS annotations generated from 5 types of distortions.

\item We propose the first LMM-based dynamic mesh quality assessment method, DynaMesh-Rater, which is a unified framework suitable for both textured 4D human mesh quality assessment and non-textured 4D human mesh quality assessment. DynaMesh-Rater integrates visual, motion, and geometry features and utilizes the LoRA-based instruction tuning technique.

\item Extensive experiment results demonstrate that our framework, DynaMesh-Rater, achieves state-of-the-art performance, which can serve as an effective objective metric for evaluating both textured and non-textured dynamic 4D digital humans.

\end{itemize}

\section{Related Work}
\label{sec:related}

\subsection{3D Quality Assessment}


\subsubsection{3D Quality Assessment Dataset}
3D mesh quality assessment datasets can be divided into static mesh quality assessment datasets and dynamic mesh quality assessment datasets.
In the early stages, research efforts are primarily directed toward artifacts in non-textured meshes, such as those arising from compression, mesh simplification, surface noise, and watermarking \cite{lavoue2006perceptually,corsini2007watermarked,torkhani2015perceptual,christaki2019subjective}. Among them, 3DMAQD \cite{torkhani2015perceptual} focuses on dynamic mesh, which contains 276 dynamic meshes generated from 10 reference non-textured meshes through 4 types of distortions, including spatial and temporal visual masking simulation, compression, and network transmission error.
Recently, several datasets for static and dynamic textured meshes have been introduced \cite{guo2016subjective,nehme2020visual,nehme2023textured,gao2025ges,zhang2023ddh,yang2024tdmd}. For instance, TMQA \cite{nehme2023textured} consists of 3000 static textured meshes generated from 55 reference meshes through 5 types of compression distortions applied to the mesh geometry, texture mapping, and texture image. In contrast to general meshes, SJTU-H3D \cite{zhang2023advancing} focuses on digital human meshes and contains 1120 meshes generated from 40 reference meshes.
However, all these datasets contain only static meshes, highlighting the need for comprehensive studies and datasets focused on the quality assessment of dynamic meshes.
In recent years, DDH-QA \cite{zhang2023ddh} and TDMD \cite{yang2024tdmd} take a significant step forward by exploring the quality assessment of dynamic meshes. Concretely, DDH-QA \cite{zhang2023ddh} is the first dynamic digital human quality assessment dataset, which contains 800 distorted dynamic meshes from only 2 reference mesh sequences and 7 distortion types. TDMD \cite{yang2024tdmd} contains 303 samples from 8 reference dynamic mesh sequences and 6 types of distortions.
However, all these datasets suffer from limitations in scale, offering too few total meshes, an insufficient number of reference meshes, and a narrow range of distortion types. Moreover, they only focus on textured meshes while neglecting the influence of degradation on mesh geometry.
To this end, we propose DHQA-4D, which consists of both textured and non-textured dynamic meshes and contains a total of 1920 meshes generated from 32 high-quality real-scanned 4D human mesh sequences and 11 diverse distortion types.



\subsubsection{3D Quality Assessment Methods}
3D quality assessment methods can be categorized into two main types: model-based methods and projection-based methods.
Model-based methods \cite{mekuria2016evaluation,tian2017geometric,alexiou2018point,torlig2018novel,yang2020graphsim,meynet2020pcqm,alexiou2020pointssim,pcqa-large-scale,zhang2022no,zhou2022blind} extract features directly from the 3D mesh, offering the benefits of viewpoint invariance and conceptual simplicity. For example, Yang \emph{et al.} \cite{yang2020graphsim} propose the GraphSIM method, which estimates perceptual point cloud quality using graph neural networks with graph signal gradients. However, such methods tend to be both resource-intensive and time-consuming due to the complexity of 3D models.

In contrast, projection-based methods \cite{wang2024zoom, yang2020predicting,chai2024plain, zhang2021mesh, fan2022no, zhang2022treating, liu2021pqa} estimate the visual quality of 3D models through their 2D projections. For example, Yang \emph{et al.} \cite{yang2020predicting} propose the first work utilizing multi-view project images and a convolutional neural network for 3D point cloud quality assessment. Wang \emph{et al.} \cite{wang2024zoom} propose the MOD-PCQA method which utilizes the multi-scale image features and propose a hierarchical multi-scale feature alignment network for static 3D point cloud quality assessment.
However, these methods are highly sensitive to viewpoint selection and may exhibit instability under different rendering configurations.

In recent years, some works \cite{chen2024dynamic, zhang2023advancing} have integrated the strengths of both model-based and projection-based methods to achieve more robust 3D quality assessment. Zhang \emph{et al.} \cite{zhang2023advancing} propose a multi-modal network that utilizes both static 3D mesh features and visual features from the vision-language model for zero-shot quality assessment of static 3D digital humans.

\subsection{LMM for Quality Assessment} 

Large multi-modality models (LMM) have become an all-powerful visual question answering model in the computer vision area. They have demonstrated strong potential on the quality assessment task \cite{ge2024lmm, wu2023q, zhang2024human,wu2023exploring, li2025aghi, wu2025fvq, zhou2026mi3s}. In recent years, numerous works have been proposed for utilizing LMM models to diverse quality assessment tasks, which include image quality assessment (IQA), video quality assessment (VQA), and point cloud quality assessment (PCQA). Among these studies, Q-bench \cite{zhang2024q} is the first work that benchmarks the inherent ability of current LMM models on predicting perceptual quality scores in a zero-shot manner. Then, Q-Align \cite{wu2023q} is the first work that utilizes instruction tuning to teach a quality assessment LMM model as a unified model for IQA and VQA tasks. In the video quality assessment area, LMM-VQA and FineVQ \cite{ge2024lmm, duan2024finevq} propose LMM models designed for user-generated content (UGC) video quality assessment by incorporating motion features into the LMM model. Recently, FVQ \cite{wu2025fvq} proposes a face-specific LMM model for evaluating the perceptual quality of in-the-wild face videos. Considering the field of 3D point cloud quality assessment, LMM-PCQA \cite{zhang2024lmm} is the first work to utilize the LMM model for evaluating the quality of static 3D point clouds. Even though there are many works utilizing Large multi-modality models for quality assessment, the works that adopt Large multi-modality models for dynamic 4D mesh quality assessment are still lacking.

\begin{figure*}[ht]
\centering
\centerline{\includegraphics[width=\linewidth]{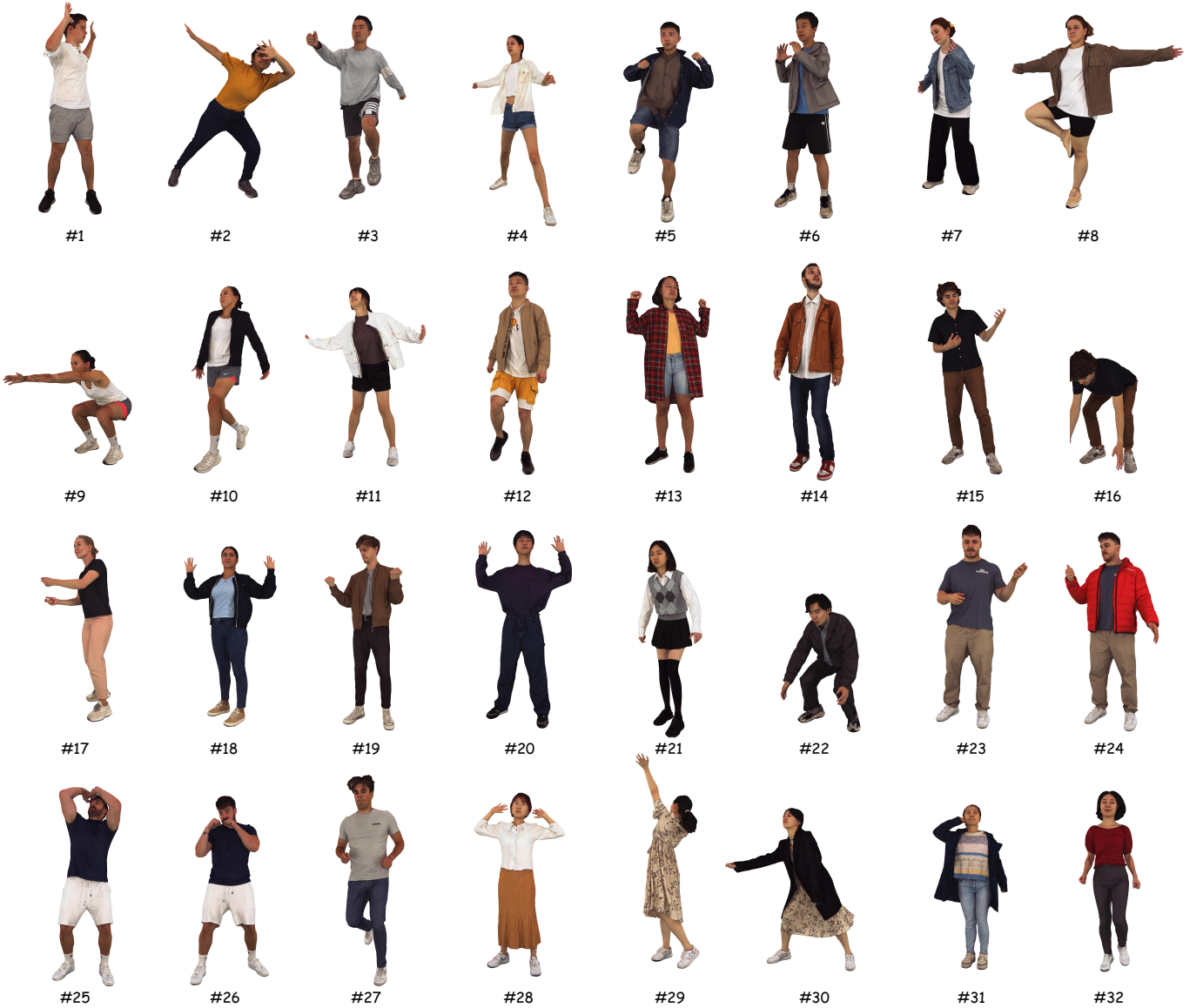}}
\caption{Snapshots of all the reference 3D digital human models in DHQA-4D dataset.}
\label{fig all_static_mesh}
\end{figure*}

\begin{figure}[ht]
\centering
\centerline{\includegraphics[width=\linewidth]{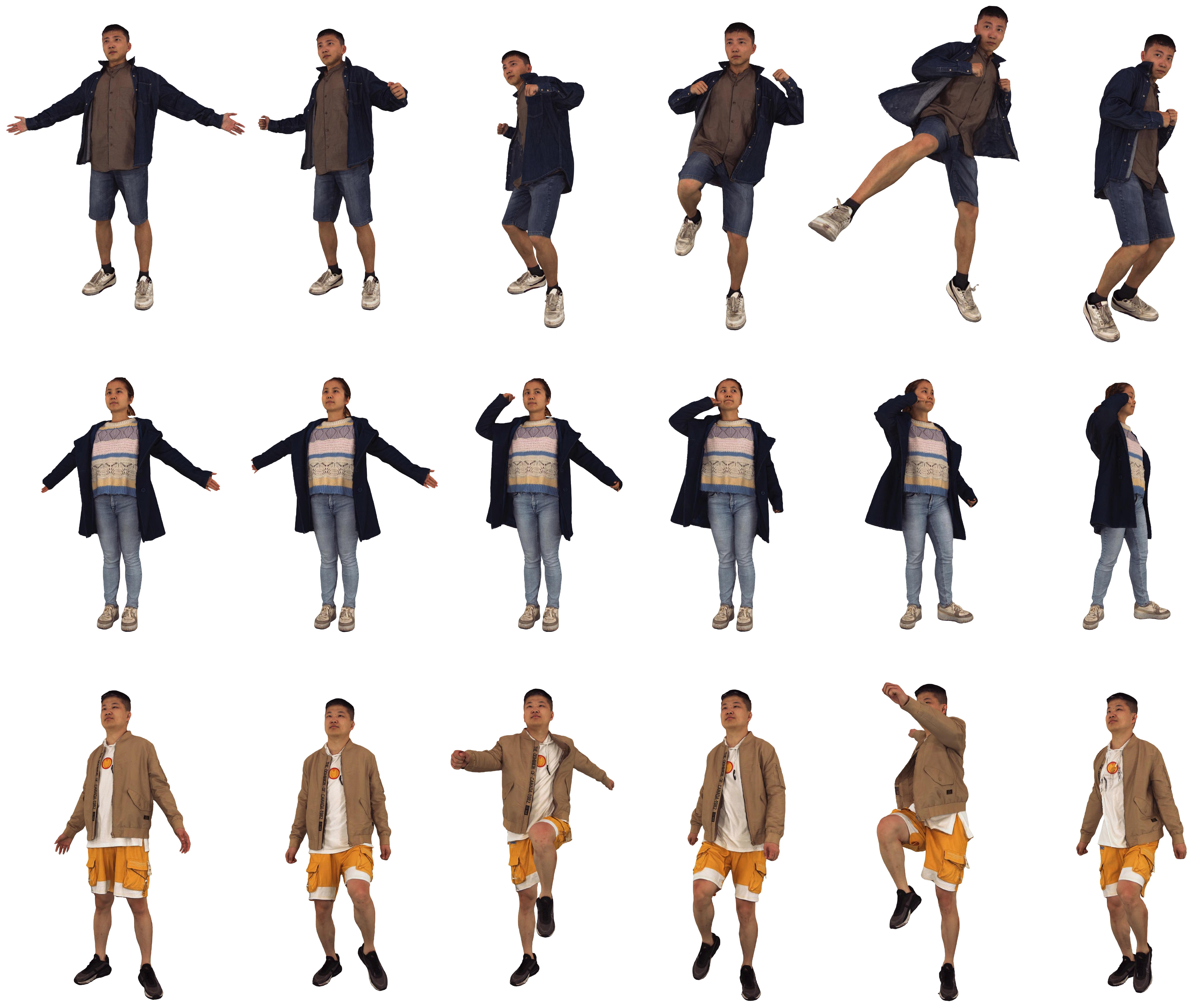}}
\caption{Visualization of example 4D mesh sequences in DHQA-4D dataset.}
\label{fig 4d_demo}
\end{figure}

\section{Dataset Construction}


\subsection{Source Data Collection}

To construct our dataset, we first carefully selected 32 high-precision dynamic human mesh data with clothing from the 4D-Dress \cite{wang20244d} dataset proposed by ETH Zurich. The detailed information of all the 4D dynamic textured scans is listed in the Table. \ref{4ddataset}. The 4D-Dress dataset is the first real-world clothing 4D human dataset containing high-quality 4D textured mesh scans. During the data collection, each participant is required to wear various types of clothing and perform a wide range of actions from 5 seconds to 8 seconds. We carefully balance the ratio of males and females, and select 32 dynamic sequences from the 4D-Dress dataset. We then crop all the sequences to 5 seconds. The visualization of all meshes is shown in Fig. \ref{fig all_static_mesh} and Fig. \ref{fig 4d_demo}. We can observe that our dataset covers a large range of human meshes in terms of appearances, actions, and clothing.

\subsection{Source Data Characterization}

In order to cover various dynamic digital human data and avoid the situation where the dataset contains too much biased data, the feature analysis of the referred digital human mesh is very important. Following the previous works \cite{zhang2023advancing, nehme2023textured}, we utilized the color feature and geometry feature to analyze the dynamic digital human meshes, considering these two features are significant for the visual characteristics of 3D digital humans.

\begin{figure}[ht]
\centering
\centerline{\includegraphics[width=\linewidth]{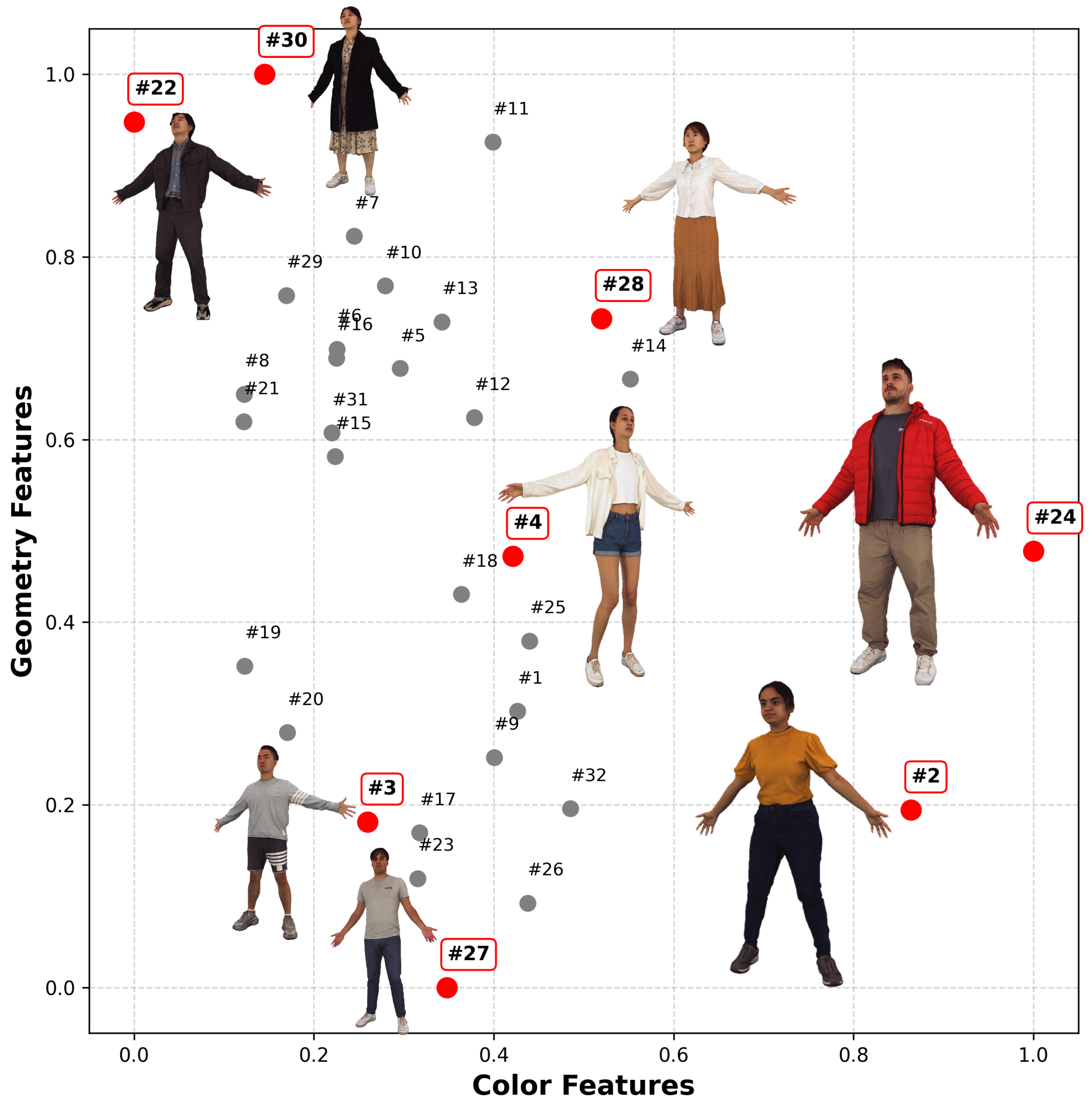}}
\caption{Visualization of geometry and color information for all reference digital human meshes.}
\label{fig:color_geometry}
\end{figure}

\subsubsection{Geometry Features}

Following previous works \cite{liu2022perceptual, nehme2023textured, zhang2023advancing}, we utilize dihedral angles to characterize the shape and curvature of the meshes. The dihedral angle denotes the angle between two neighboring faces that share an edge within the mesh, which conveys valuable insights about the smoothness and sharpness of the mesh surfaces. To compute the overall geometry feature of each dynamic human mesh, we calculate the standard deviation of all the dihedral angles in each mesh:
\begin{equation}
F_{G} = std(Mesh_{Dihedral}),
\end{equation}
where $F_{G}$ denotes the geometry feature and $std()$ denotes the calculation of standard deviation. With the help of geometry features, we aim to assess the geometric characteristics of each dynamic digital human mesh. We take the average of the geometric features of the mesh of all frames as the geometric features of each dynamic human mesh.

\subsubsection{Color Features} We extract color features from the texture map of each human mesh. Following previous works \cite{fairchild2013color, hasler2003measuring}, we begin by converting the texture from RGB to LAB color space. Then, we compute the combined standard deviation of the A and B channels, which is represented using the following equation:
\begin{equation}
F_{C} = \sqrt{std(A)^2+std(B)^2},
\end{equation}
where $F_{C}$ denotes the colorful feature, $A$ and $B$ denote the corresponding color channel features from the texture map. We take the average of the color features of the mesh of all frames as the color features of each dynamic human mesh.

\subsubsection{Characterization Visualization} We report the joint color and geometry analysis results of all 32 dynamic digital human meshes in Fig. \ref{fig:color_geometry}. The figure demonstrates that our reference dynamic human meshes cover a wide range of geometry features and color features. To be noticed, we can observe that model $\#24$ placed on the right has the most prominent colors. The model $\#30$ placed on the top-left corner shows a monotonous dark color, and it has rich geometric details because the digital human is wearing outerwear. Meanwhile, the model $\#3$ placed on the bottom-left corner contains simple geometry information and relatively subdued colors.

\subsection{Distortion Data Creation}

As demonstrated in the Table \ref{4ddataset}, different from previous works containing limited distortions, we comprehensively consider potential distortions which may affect the visual quality of dynamic meshes and divide them into 12 subsets: geometrical gaussian noise (GN), color noise (CN), texture downsampling (TD),  mesh simplification (MS), texture map compression (TMC), position compression (PC), UV map compression (UMC), position and UV map compression (PUC), Geometry and Texture Compression (GTC), mixed compression (MC) and temporal discontinuity (DC).

For each distortion, we manually select proper distortion parameters to cover most visual quality scope, and the detailed distortions are introduced as follows:
\begin{itemize}
\item Geometrical Gaussian Noise (GN): Geometrical Gaussian Noise is added to the vertex coordinates of the dynamic human meshes following \cite{yang2024tdmd}. The human mesh is distorted with seven levels, including 0.001, 0.004, 0.007, 0.010, 0.013, 0.016, and 0.020.

\item Color Noise (CN): The salt pepper noises with seven noise levels 0.01, 0.05, 0.1, 0.15, 0.2, 0.25, and 0.4, which are generated by matlab function ``imnoise" are applied to the texture maps of dynamic meshes.

\item Texture Downsampling (TD): We utilize the ``imresize" function in matlab to process the original texture maps with a resolution of $1024 \times 1024$ to resolutions of $512 \times 512$, $308 \times 308$, $205 \times 205$, $103 \times 103$, $82 \times 82$, $52 \times 52$ and $31 \times 31$, respectively.

\begin{figure}[ht]
\centering
\centerline{\includegraphics[width=\linewidth]{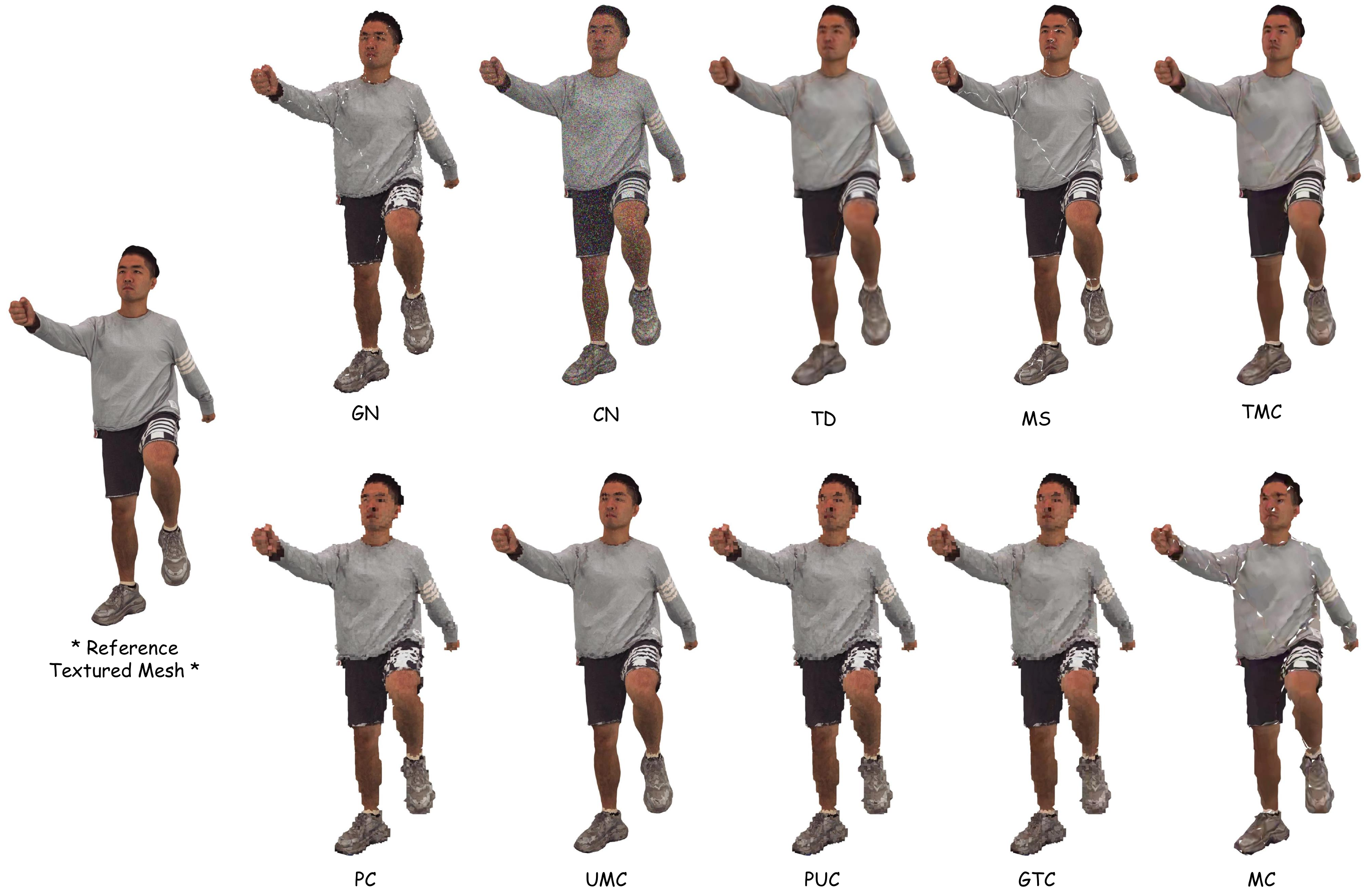}}
\caption{Visualization of distorted textured meshes in DHQA-4D dataset. The DC distortion type is ignored, considering this type does not bring distortions to geometry and texture.}
\label{fig}
\end{figure}

\begin{figure}[ht]
\centering
\centerline{\includegraphics[width=\linewidth]{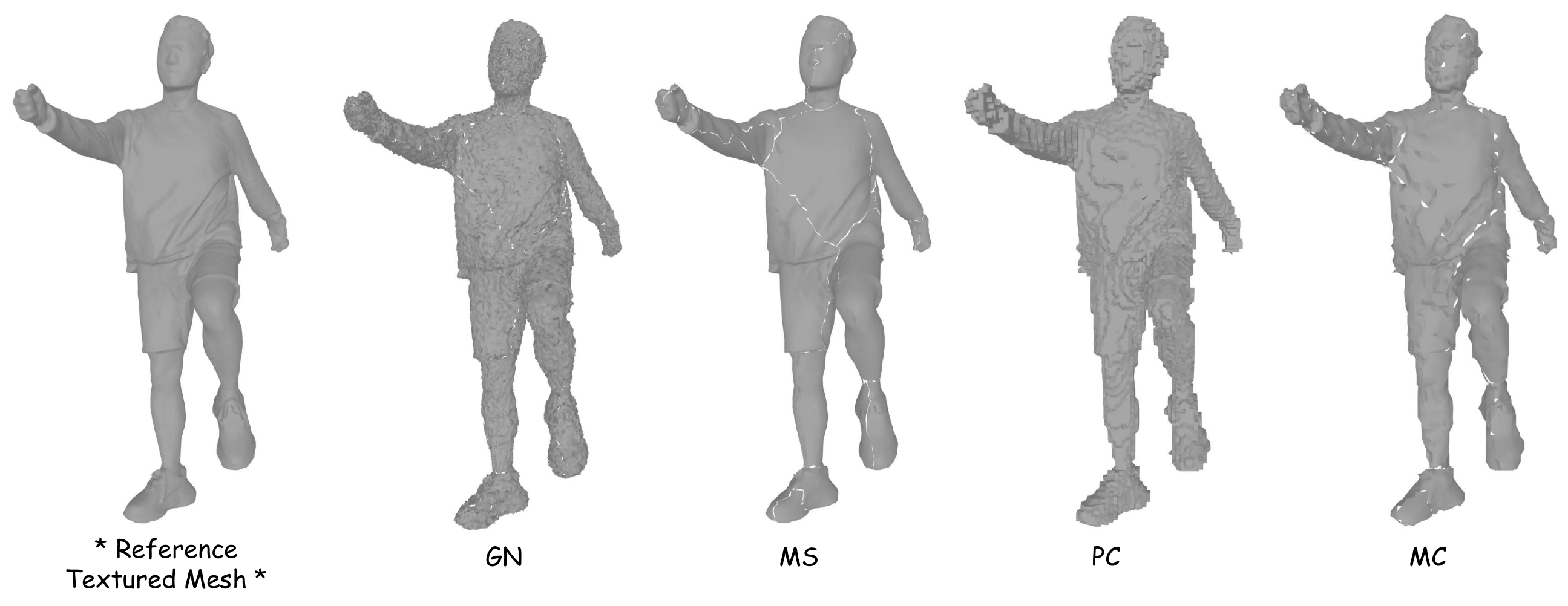}}
\caption{Visualization of distorted non-textured meshes. Similarly, the DC distortion type is ignored.}
\label{fig}
\end{figure}

\item Mesh Simplification (MS): Mesh simplification is an effective method to compress the size of mesh files through reducing the number of vertices and faces, making the model contain different levels of detail. Following \cite{yang2024tdmd}, we utilize the quadric edge collapse decimation in PyMeshLab to simplify the 3D dynamic mesh. We specifically set the simplified faces number $F_n$ to be 5K, 10K, 20k, 30k, 40k.

\item Texture Map Compression (TMC): We apply the popular FFMPEG 5.1.2 algorithm to compress the texture map images to evaluate the visual quality influence of image compression methods. Concretely, we select the quantization parameters $T_q$ to be 22, 37, 48, and 50.

\item Position Compression (PC): We utilize the Draco library to quantize the position information of each 4D digital human mesh. The compression parameter $Q_p$ is selected as 6, 7, 8, 9. 

\item UV Map Compression (UMC): Similar to position compression, we utilize the Draco library to quantize the texture coordinate information with specific compression parameter $Q_t$ to be 7, 8, 9, 10.

\item Position and UV Map Compression (PUC): Instead of compressing dynamic meshes with a single compression parameter, we also define several combinations of $Q_p$ and $Q_t$ parameters to simulate a complex mesh compression process during transmission. We select (6, 7), (7, 7), (8, 8), (9, 9), and (9, 10) as our parameter combinations.

\item Geometry and Texture Compression (GTC): We comprehensively consider a hybrid compression scenario that involves using Draco for mesh compression and FFmpeg for image compression simultaneously. Hence, we define the combination of $(Q_p, Q_t, T_q)$ to be (6, 7, 50), (7, 7, 37), (8, 8, 37), (9, 9, 37), and (9, 9, 22) respectively.

\item Mixed Compression (MC): Mixed compression considers practical applications where mesh simplification, Draco compression, and texture map compression are applied simultaneously. We define the combination of $(Q_p, Q_t, F_n, T_q)$ to be (6, 7, 5K, 50), (7, 7, 5K, 37), (7, 8, 10K, 50), (8, 8, 10K, 37), (8, 10, 10K, 22), (9, 9, 20K, 37), (9, 10, 20K, 22) and (9, 10, 40K, 22) respectively.

\item Temporal Discontinuity (DC): Previous research \cite{zhou2024subjective} has found that the stuck phenomenon or frame drop phenomenon during the video transmission process will produce an unpleasant quality experience. Therefore, considering these temporal distortions for a dynamic 4D mesh is also important. Specifically, we introduce a time stuck parameter $T_s$ to simulate the time (expressed in seconds) when the dynamic mesh gets stuck during transmission. We also introduce a frame drop parameter $T_d$ to simulate the frame loss time (expressed in seconds) of the dynamic mesh during the transmission process. We set the $T_s$ and $T_d$ to be (1,2) and (1,2). The stuck and drop events are all introduced in after the first 2 seconds.

\end{itemize}

After this process, we end up with $1920=32\times(7\times4+10+4*2+5*2+4)$ distorted textured mesh sequences. Among these textured meshes, we also extracted $832=32\times(7+4+4+5+6)$ non-textured meshes with different geometry distortions, including GN, MS, PC, MC, and TC, where we remove the redundant parameters in MC distortion because some combination such as (8,8,10k,37) and (8,10,10k,22) will generate similar distortion results for non-textured meshes, resulting in 6 unique distortions in MC mode.

\begin{figure*}[ht]
\vspace{-1mm}
\centering
\centerline{\includegraphics[width=0.95\linewidth]{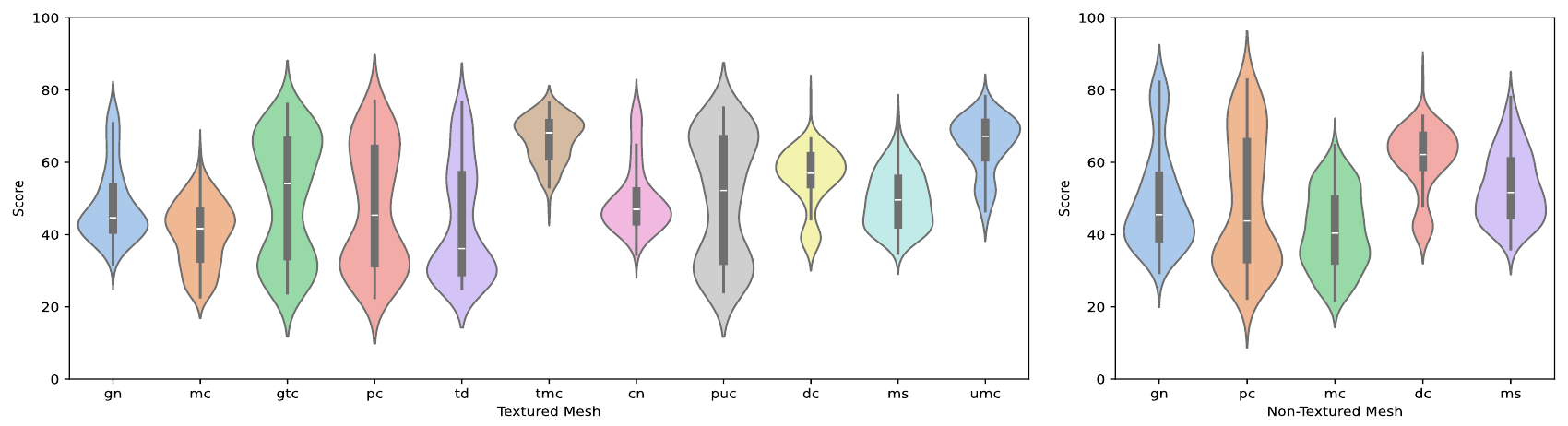}}
\vspace{-1mm}
\caption{The MOS distributions in terms of different distortion types in textured 4D mesh subset and non-textured 4D mesh subset in DHQA-4D dataset.}
\label{fig_violin}
\end{figure*}

\begin{figure}[ht]
\centering
\centerline{\includegraphics[width=\linewidth]{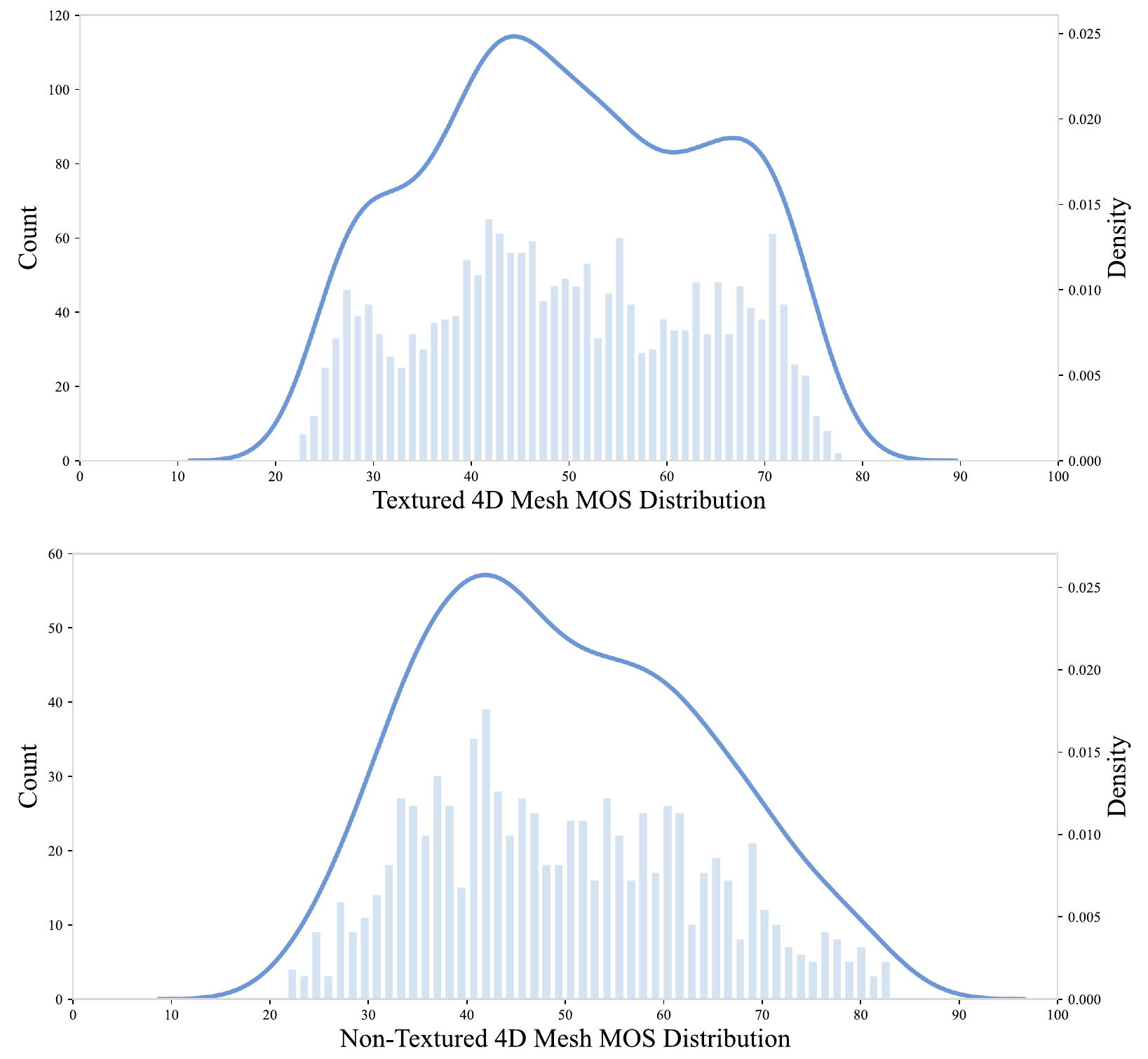}}
\caption{Distribution of MOSs of textured 4D meshes and non-textured 4D meshes in DHQA-4D dataset.}
\label{fig_overall_mos}
\end{figure}

\subsection{Subjective Experiment}

A large-scale subjective experiment is conducted where 25 participants are recruited to score the DHQA-4D dataset. We detailedly explain our subjective experiment setting from the perspectives of 4D mesh rendering, experiment setup, and subjective data processing.

\subsubsection{4D Mesh Rendering}

To comprehensively assess the perceptual quality of a dynamic 4D digital human, rendering the 4D mesh in a fixed camera view as in \cite{zhang2023ddh} is not reasonable because the observer can not see the human's backside, which may impact the perceptual visual quality assessment of the overall dynamic mesh. Therefore, we position the first frame of the dynamic mesh at the origin of the 3D coordinate system. Subsequently, we utilize a dynamic camera that rotates 360 degrees horizontally around the dynamic human mesh to enhance the visual assessment experience. Specifically, for a 5-second motion sequence, we first position the digital human facing the camera, with the camera remaining stationary for one second. Then, over the next four seconds, the camera rotates 360 degrees horizontally, returning to its initial position. We found that this rendering approach effectively allows for a comprehensive observation of the dynamic digital human from multiple perspectives.

\subsubsection{Participants and Experiment Setup}

To obtain the perceptual quality score of each dynamic human mesh, we invite 25 subjects and conduct the subjective experiment under the guidance in ITU-R BT.500 \cite{series2012methodology}. We carefully prepare an indoor experimental environment with standard lighting, and the dynamic 3D digital human is rendered on a 1920*1080 resolution screen. The subjects are asked to rate the perceptual score on a scale of 0 to 5, where 0 denotes the worst quality and 5 denotes the best quality. All the subjects are guided by a detailed rating explanation and several testing samples before the formal experiments. Considering that the data set is very large, we split the 1920 videos into 6 equal groups, each of which contains 320 videos. During the experiment, each group lasts approximately 1 hour, then the subjects are asked to have a 20-minute break before turning to the next group.


\subsubsection{Subjective Data Processing}

After achieving raw scores for each subject, we carefully follow the ITU guidelines \cite{series2012methodology} to analyze and process our collected subjective perceptual quality scores. Concretely, we first identify the outliers of all collected scores and disqualify unreliable participants. For all the perceptual scores of each distorted dynamic human mesh, we compute the kurtosis of the raw scores assigned to each dynamic human mesh to determine if the data follows a Gaussian or non-Gaussian distribution. For the Gaussian distribution, we identify the outlier score when the score deviates by more than 2 standard deviations (std) from the mean score. For the non-Gaussian distribution, we identify the outlier score when the score deviates by more than $\sqrt{20}$ standard deviations (std) from the mean score. We remove the participants with an outlier rate surpassing $5\%$ from our dataset. After this process, we conduct z-score normalization and transform the filtered raw scores into Z-scores, which are normalized to a range of [0, 100]. The processed mean opinion score (MOS) is calculated as follows:
$$z_i{}_j=\frac{r_i{}_j-\mu_i{}_j}{\sigma_i},\quad z_{ij}'=\frac{100(z_{ij}+3)}{6},$$
$$\mu_i=\frac{1}{N_i}\sum_{j=1}^{N_i}r_i{}_j, ~~ \sigma_i=\sqrt{\frac{1}{N_i-1}\sum_{j=1}^{N_i}{(r_i{}_j-\mu_i{}_j)^2}}$$ 
where $r_{ij}$ denotes the raw ratings annotated by the $i$-th subject according to the $j$-th video. $N_i$ is the number of videos annotated by subject $i$. Finally, the MOS of the $j$-th video is calculated by averaging the rescaled z-scores through the following function: $$MOS_j=\frac{1}{M}\sum_{i=1}^{M}z_{ij}'$$
where $MOS_j$ represents the MOS for the $j$-th dynamic 4D digital human mesh, $M$ is the number of subjects, and $z'_i{}_j$ are the rescaled z-score.

\subsection{Dataset Statistics and Analysis}

\subsubsection{Overall  MOS Observations.} The overall MOS score distribution is shown in Fig. \ref{fig_overall_mos}. We can observe that both the MOSs for textured 4D meshes and non-textured 4D meshes span wide score distributions. We can observe that the MOS distribution of the textured mesh is more uniform. The reason might be that the textured meshes subset contains more types of distortion noise, including geometry distortions, texture distortions, and the combinations of these two distortions.

\subsubsection{Influence of Single Distortion on Perceptual Quality} In addition to the analysis of overall MOS, we further analyze the score distribution of each distortion type in both the textured mesh subset and the non-textured mesh subset. The visualization results are summarized in Fig. \ref{fig_violin}. First, we can observe that for both subsets, the GTC, PC, TD, and PUC distortions cover a wide range of MOSs. The scores are all ranged from 10 to 90. Second, we can observe that for the TMC and UMC distortions, the MOS scores are pretty high. It demonstrates that the UMC distortions on texture coordinate bring less visual impact on the quality of experience, and our selected compression parameters for TMC contain few distortions. Moreover, we can observe that the DC distortion has relatively high MOS scores in both subsets, demonstrating that the distortions in the temporal domain do not largely affect the visual experience compared to the texture distortions and geometry distortions.


\section{DynaMesh-Rater Method}

\begin{figure*}[ht]
\centering
\centerline{\includegraphics[width=0.98\linewidth]{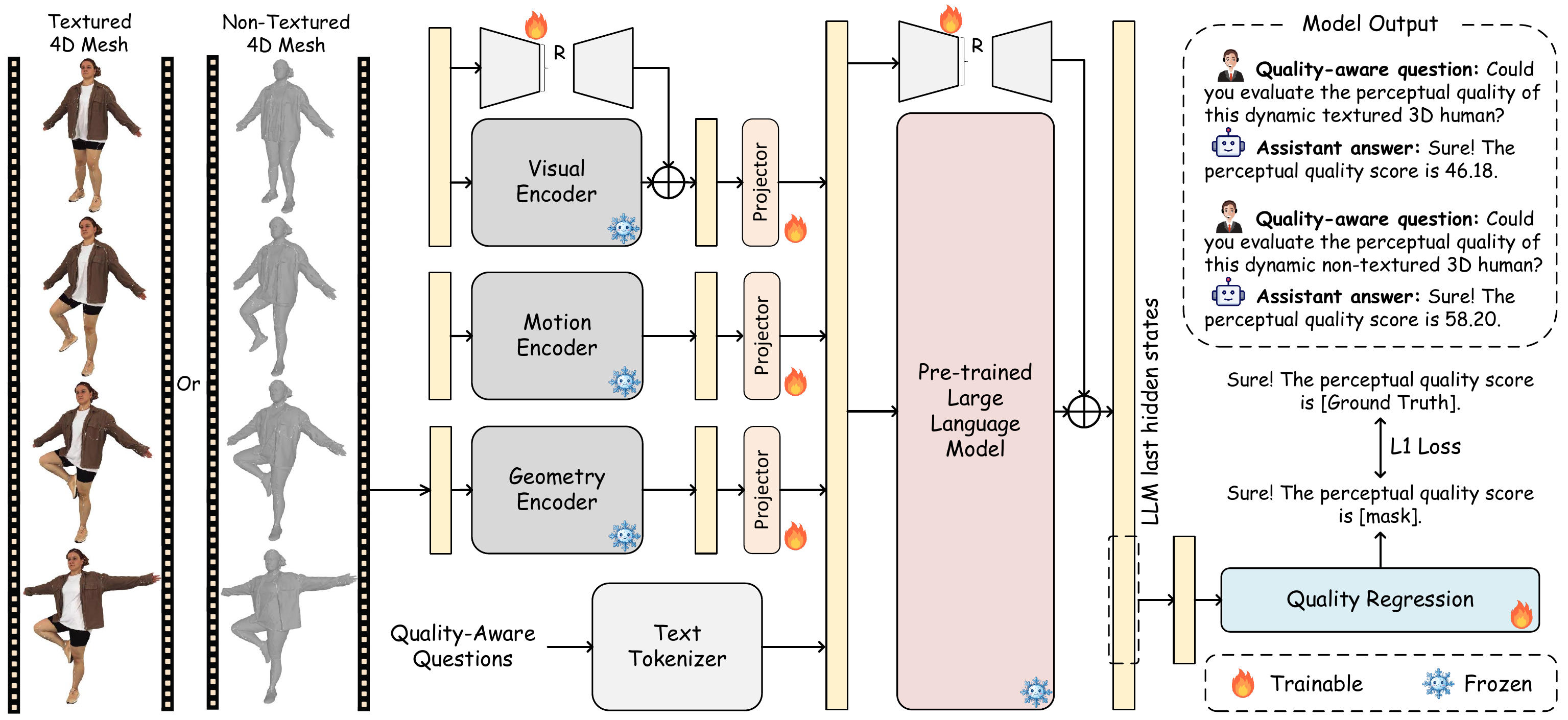}}
\caption{The framework of our proposed DynaMesh-Rater. DynaMesh-Rater includes four feature encoders, including a vision encoder to extract visual features from sparse projected video frames, a motion encoder to extract motion features from uniformly cropped video clips, a geometry encoder to extract shape-aware features from the surfaces of dynamic human meshes, and a text encoder to extract language features from user questions. Then all the extracted features are aligned by corresponding projectors and fed into a large language model for regressing quality scores.}
\label{fig_method}
\end{figure*}

Along with our proposed DHQA-4D dataset, we further propose \textbf{DynaMesh-Rater}, a novel LMM-based approach especially designed for dynamic 4D mesh quality assessment. The whole framework is demonstrated in Fig. \ref{fig_method}.

\subsection{Problem Definition} We first briefly introduce the quality assessment task and our framework. Given a 3D human mesh sequence $M = (m_1, m_2, ..., m_{T})$ where $T$ denotes the sequence length and a projected video $V=(v_1, v_2, ..., v_{T})$ rendered from the mesh sequence $M$, the quality assessment task aims to regress a quality score $S\in R^1$ where $S$ is a contiguous number from 0 to 100.

\subsection{Overall Architecture} DynaMesh-Rater consists of four feature encoders, including a vision encoder to extract visual features from sparse projected video frames, a motion encoder to extract motion features from uniformly cropped video clips, a geometry encoder to extract shape-aware features from the surfaces of dynamic human meshes, and a text encoder to extract language features from user questions. Then all the extracted features are aligned by corresponding projectors and fed into a large language model for regressing quality scores.

\subsection{Model Design}

\subsubsection{Visual Features Encoding} As demonstrated in Fig. \ref{fig_method}, considering that current LMM models can only cope with a limited number of images, we first convert the projected digital human video into sparse image frames $V_f$. We select a modern vision transformer (InternViT \cite{chen2024internvl}) trained with contrastive text-to-image learning to extract the corresponding visual features. To ensure compatibility between the extracted visual features and the input dimensions of the large language model (LLM), we employ two MLP layers as a projector layer to transform the visual features. This projection process is formally expressed as:
\begin{equation}
F_v = \mathcal{P}_V(\mathcal{E}_I(V_f)),
\end{equation}
where $F_v$ is the projected visual features compatible with the LLM input space.

\subsubsection{Motion Features Encoding} In addition to the extracted static visual features from sparse image frames, we also extract motion features from the projected video. Concretely, to comprehensively capture the whole dynamics of human motions, we first uniformly crop the whole projected video into five video clips $(C_1, ..., C_5)$. Then we utilize a pre-trained SlowFast model \cite{feichtenhofer2019slowfast} as the motion encoder $\mathcal{E}_M$ to extract motion features from the video $(C_1, ..., C_5)$. We then utilize a two-layer MLP (Multilayer Perception) network $\mathcal{P}_M$ to project the motion features to the input space of LLM:
\begin{equation}
F_m = \mathcal{P}_M(\mathcal{E}_M(C_1, ..., C_5)),
\end{equation}
where $F_m$ is the projected motion features compatible with the LLM input space.


\subsubsection{Geometry Features Encoding} Relying exclusively on visual input for dynamic mesh quality evaluation may neglect the role of geometric information in perceptual quality, ultimately decreasing the performance of the assessment results. Hence, we extract the geometry features from the 4D meshes and incorporate them into our LMM-based framework. A previous study has demonstrated the usefulness of geometry attributes in assessing mesh quality. In our paper, we select the dihedral angle as the basic mesh feature because it is invariant to scale compared to other geometry attributes such as curvature. The dihedral angle $\theta$ denotes the angle between two adjacent mesh faces, which is computed as the dot product of two corresponding normal vectors. After obtaining the dihedral angle $\theta$ for all adjacent mesh faces, we extract the overall geometry features $O_g$ by calculating diverse distribution parameters of all dihedral angles. Concretely, $O_g$ is usually denoted by
\begin{equation}
O_g = ( Basic(\theta), GGD(\theta), Gamma(\theta) ),
\end{equation}
where the $Basic()$ function aims to calculate the mean, variance, and entropy of all dihedral angles. The $GGD$ function represents calculating the generalized Gaussian distribution (GGD) parameters of all dihedral angles. The $Gamma$ function calculates the shape parameters and scale parameters of the Gamma distribution for all dihedral angles. After achieving the $O_g$ feature, we finally utilize a two-layer MLP network $\mathcal{P}_G$ to map geometry features $O_g$ to the input space of LLM:
\begin{equation}
F_g = \mathcal{P}_G( O_g ),
\end{equation}
where $F_g$ is the projected geometry features compatible with the LLM input space.

\subsubsection{Quality Regression} After obtaining the multidimensional features $F_v$, $F_m$, and $F_g$, all the features are concatenated together and fed into the large language model (LLM) model to predict the quality score from the output text head of the LMM. Different from previous research \cite{wu2023q}, which transforms predicted MOS scores into 5 discrete text-defined levels (bad, poor, fair, good, and excellent), we utilize an MLP network to directly regress continuous scores from the last hidden states of the LMM model. Through predicting numerical scores instead of discrete quality levels, our DynaMesh-Rater can flexibly fit the continuous perceptual quality scores.


\section{Experiments}

\subsection{Experimental Setup}

\subsubsection{Benchmark Dataset} Considering that our dataset is the first multidimensional quality assessment dataset that includes perceptual quality scores on both textured dynamic human meshes and non-textured dynamic human meshes. We split the DHQA-4D dataset into a textured mesh subset and a non-textured mesh subset. The textured mesh subset contains 1916 distorted textured 4D mesh sequences and 832 non-textured 4D mesh sequences.

\begin{table}[h]
\centering
\renewcommand\arraystretch{1.4}
\caption{ \textbf{Performance comparisons} on the DHQA-4D dataset. $\heartsuit$, $\clubsuit$, and $\spadesuit$ denote full reference quality assessment method, no-reference image quality assessment method, and no-reference video quality assessment method, respectively. The best performances are bold.}
\vspace{-1mm}

\resizebox{1.01\linewidth}{!}{
\begin{tabular}{lcccccc}
\toprule
Dataset  & \multicolumn{3}{c}{Textured Mesh}  & \multicolumn{3}{c}{Non-Textured Mesh}  \\ \cmidrule(lr){2-4} \cmidrule(lr){5-7} 

Method / Metric  & SRCC\,$\uparrow$  & PLCC\,$\uparrow$  & KRCC\,$\uparrow$  & SRCC\,$\uparrow$  & PLCC\,$\uparrow$  & KRCC\,$\uparrow$ \\ 
\midrule

$\heartsuit$ PSNR$_{rgb}$  & 0.6473	& 0.6492 &	0.4759 &  0.7094	& 0.6835 & 0.5074	\\
$\heartsuit$ PSNR$_{yuv}$  &  0.6380 & 0.6481 & 0.4683 & 0.7097 & 0.6840 & 0.5073\\

$\heartsuit$ SSIM \cite{wang2004image}	& 0.6869 &	0.4902	& 0.4998 & 0.5321 &	0.4223	& 0.4025 \\
$\heartsuit$ MS-SSIM \cite{wang2003multiscale}	&  0.6636 &	0.4211 & 0.4797 & 0.4784 & 0.3068	& 0.3662 \\
$\heartsuit$ GMSD \cite{xue2013gradient}	& 0.7028 & 0.6806	& 0.5131 &	0.7036	& 0.6819 &	0.5067 \\
$\heartsuit$ G-LPIPS \cite{nehme2023textured}	& 0.7262 & 0.7220 & 0.5344 &	0.4388 & 0.6031 &	0.3591 \\

\hdashline

$\clubsuit$ HyperNet \cite{su2020blindly}	& 0.7076 &  0.7071 & 0.5291 & 0.8120 & 0.7765 & 0.6256 \\

$\clubsuit$ MUSIQ \cite{ke2021musiq} & 0.7731 & 0.7649 & 0.6412 & 0.8565 & 0.8495 & 0.6873 \\

$\clubsuit$ MANIQA \cite{yang2022maniqa} & 0.8772 & 0.8872 & 0.7019 & 0.8906 & 0.8809 & 0.7193 \\

\hdashline

$\spadesuit$ VSFA \cite{li2019quality} & 0.8782 & 0.8736 & 0.6946 & 0.8648 & 0.8690 & 0.7183 \\

$\spadesuit$ GSTVQA \cite{chen2021learning}	& 0.8928  & 0.9002 & 0.7315 & 0.8863	& 0.8912 &	0.7276 \\

$\spadesuit$ SimpleVQA \cite{sun2022deep}	&  0.8842 &	0.8937 & 0.7105 & 0.8921 & 0.8793 & 0.7261 \\

$\spadesuit$ FastVQA \cite{wu2022fast}	&	0.8585 & 0.8467 & 0.6752 & 0.8916 & 0.8699 & 0.7222 \\

$\spadesuit$ Dover \cite{wu2023exploring}	& 	0.8693 & 0.8617 & 0.6904 & 0.8969 & 0.8829 & 0.7336 \\

$\spadesuit$ KSVQE \cite{lu2024kvq}	&	0.9043 & 	0.8978	& 0.7359 & 0.9134 &	0.8961 & 0.7566 \\

\hdashline
\rowcolor[gray]{.92}
\textbf{DynaMesh-Rater}	& \textbf{0.9316} & \textbf{0.9327} & \textbf{0.7762} & \textbf{0.9217} & \textbf{0.9161} & \textbf{0.7713}
\\

\bottomrule
\end{tabular}
}
\vspace{-1mm}
\label{tab:exp_overall_mos}
\end{table}

\subsubsection{K-fold Cross-Validation} Following the previous works \cite{zhang2023ddh, liu2022perceptual}, we adopt a $k$-fold cross-validation strategy to perform robust evaluations. We first divide the data set into $k$ data parts of equal size in terms of human identity. Considering that the DHQA-4D dataset contains 32 reference 4D meshes, we select a value of $k = 4$ for the k-fold cross-validation. We carefully split the dataset to make sure that there is no data overlap between the training and testing folds.

\begin{table*}[h]
\centering
\renewcommand\arraystretch{1.7}
\caption{ \textbf{Benchmark performance} on the textured mesh subset of the DHQA-4D dataset for different distortion types. $\heartsuit$, $\clubsuit$, and $\spadesuit$ denote full reference quality assessment methods, no-reference image quality assessment methods, and no-reference video quality assessment methods, respectively.}
\resizebox{\linewidth}{!}{
\begin{tabular}{l|*{11}{cc}}
\toprule
\multirow{2}{*}{\makecell{\textbf{Method}}} 
& \multicolumn{2}{c}{\textbf{GN}} & \multicolumn{2}{c}{\textbf{CN}} & \multicolumn{2}{c}{\textbf{TD}}
& \multicolumn{2}{c}{\textbf{MS}} & \multicolumn{2}{c}{\textbf{TMC}} & \multicolumn{2}{c}{\textbf{PC}}
& \multicolumn{2}{c}{\textbf{UMC}} & \multicolumn{2}{c}{\textbf{PUC}} & \multicolumn{2}{c}{\textbf{GTC}}
& \multicolumn{2}{c}{\textbf{MC}} & \multicolumn{2}{c}{\textbf{DC}} \\
\cmidrule(lr){2-3} \cmidrule(lr){4-5} \cmidrule(lr){6-7} \cmidrule(lr){8-9}
\cmidrule(lr){10-11} \cmidrule(lr){12-13} \cmidrule(lr){14-15} \cmidrule(lr){16-17}
\cmidrule(lr){18-19} \cmidrule(lr){20-21} \cmidrule(lr){22-23}
& SRCC$\uparrow$ & PLCC$\uparrow$
& SRCC$\uparrow$ & PLCC$\uparrow$
& SRCC$\uparrow$ & PLCC$\uparrow$
& SRCC$\uparrow$ & PLCC$\uparrow$
& SRCC$\uparrow$ & PLCC$\uparrow$
& SRCC$\uparrow$ & PLCC$\uparrow$
& SRCC$\uparrow$ & PLCC$\uparrow$
& SRCC$\uparrow$ & PLCC$\uparrow$
& SRCC$\uparrow$ & PLCC$\uparrow$
& SRCC$\uparrow$ & PLCC$\uparrow$
& SRCC$\uparrow$ & PLCC$\uparrow$ \\
\midrule

$\heartsuit$ PSNR$_{rgb}$	&	0.8269 	&	0.9160 	&	0.8323 	&	0.9220 	&	0.8763 	&	0.9245 	&	0.8979 	&	0.8958 	&	0.6529 	&	0.7001 	&	0.8978 	&	0.8962 	&	0.7130 	&	0.7834 	&	0.8531 	&	0.9139 	&	0.8854 	&	0.9220 	&	0.8871 	&	0.8930 	&	0.5295 	&	0.5214 \\
$\heartsuit$ PSNR$_{yuv}$	&	0.8264 	&	0.9163 	&	0.8232 	&	0.9207 	&	0.8694 	&	0.9151 	&	0.8985 	&	0.8971 	&	0.6023 	&	0.7000 	&	0.8909 	&	0.8936 	&	0.6877 	&	0.7818 	&	0.8506 	&	0.9118 	&	0.8774 	&	0.9198 	&	0.8861 	&	0.8920 	&	0.3864 	&	0.5218 \\
$\heartsuit$ SSIM \cite{wang2004image}	&	0.8410 	&	0.7466 	&	0.8034 	&	0.7622 	&	0.9150 	&	0.8078 	&	0.9024 	&	0.8011 	&	0.5696 	&	0.6589 	&	0.8977 	&	0.8614 	&	0.7254 	&	0.8030 	&	0.8775 	&	0.8897 	&	0.9068 	&	0.8981 	&	0.9185 	&	0.9000 	&	0.3224 	&	0.2032 \\
$\heartsuit$ MS-SSIM \cite{wang2003multiscale}	&	0.8364 	&	0.7275 	&	0.8179 	&	0.7992 	&	0.8988 	&	0.7974 	&	0.9032 	&	0.7966 	&	0.5817 	&	0.6666 	&	0.8919 	&	0.8559 	&	0.7392 	&	0.8052 	&	0.8889 	&	0.8892 	&	0.9023 	&	0.8946 	&	0.9250 	&	0.8991 	&	0.3175 	&	0.1963 \\
$\heartsuit$ GMSD \cite{xue2013gradient}	&	0.8283 	&	0.8569 	&	0.8235 	&	0.8417 	&	0.8749 	&	0.8619 	&	0.8924 	&	0.8586 	&	0.6847 	&	0.7243 	&	0.8993 	&	0.8899 	&	0.7620 	&	0.8221 	&	0.8685 	&	0.9040 	&	0.9013 	&	0.9184 	&	0.9039 	&	0.9024 	&	0.4271 	&	0.2362 \\
 $\heartsuit$ G-LPIPS \cite{nehme2023textured}	&	0.8247 	&	0.9092 	&	0.7967 	&	0.8761 	&	0.8735 	&	0.9269 	&	0.9007 	&	0.8996 	&	0.5897 	&	0.6855 	&	0.9273 	&	0.9362 	&	0.6517 	&	0.7947 	&	0.8684 	&	0.9384 	&	0.8939 	&	0.9423 	&	0.8304 	&	0.8202 	&	0.3999 	&	0.4765 \\

\hdashline

$\clubsuit$ HyperNet \cite{su2020blindly}& 0.5437  & 0.5799  & 0.4537  & 0.6016  & 0.7475  & 0.7816  &  0.6397 & 0.5493  & 0.2919  & 0.2605  & 0.7333  & 0.7183  & 0.1888  & 0.1582  & 0.7126  & 0.7511  & 0.7474  & 0.7571  & 0.7471  & 0.7199  & 0.1312  & 0.0891  \\
$\clubsuit$ MUSIQ \cite{ke2021musiq} & 0.7836  & 0.8055  & 0.6724  & 0.6913  & 0.8192  & 0.8445  & 0.8421   & 0.8409  & 0.3786  & 0.3850  & 0.8538  & 0.8474  & 0.4355  & 0.4859  & 0.8571  & 0.8644  & 0.8372  & 0.8610  &  0.8446 & 0.8462  & 0.0492  & 0.0533  \\
$\clubsuit$ MANIQA \cite{yang2022maniqa}	& 0.8102 & 0.8862 & 0.7292 & 0.8487 & 0.8622 & 0.9250 & 0.8521 & 0.8296 & 0.4930 & 0.5018 & 0.8716 & 0.8821 & 0.4712 & 0.5723 & 0.8909 & 0.9451 & 0.9034 & 0.9464 & 0.8795 & 0.8786 & 0.0347 & 0.0472 \\

\hdashline

$\spadesuit$ VSFA \cite{li2019quality}	& 0.7748	& 0.8417  &  0.7566	& 0.8723 &  0.8508 & 0.8939 &  0.8086	& 0.8391 &   0.4852	& 0.5657 & 0.9031	 &  0.9378 &  0.5540 & 0.6439 & 0.8657 & 0.9283 &  0.8826	& 0.9328 &  0.8356	& 0.8471 &  0.0872  & 0.0961  \\
$\spadesuit$ GSTVQA \cite{chen2021learning}	&	0.8038 	& 0.8591  & 0.7427  & 0.8264  & 0.8692  & 0.9045  & 0.8170  & 0.8194  & 0.4740  & 0.4981  & 0.9164  & 0.9278  & 0.6057  & 0.6466  & 0.8683  & 0.8979  & 0.9003  & 0.9284 & 0.8455  & 0.8596  &  0.1568  &  0.0862  \\
$\spadesuit$ SimpleVQA \cite{sun2022deep}	& 0.7920	& 0.8857  &  0.7217	& 0.8623 &  0.8703 & 0.9343 &  0.8238	& 0.8169 &   0.4485	& 0.5238 & 0.9143	 &  0.9532 &  0.6153 & 0.7017 & 0.8928 & 0.9598 &  0.9038	& 0.9566 &  0.8627	& 0.8647 &  0.1446	& 0.2163 \\

$\spadesuit$ FastVQA \cite{wu2022fast}	&	0.8122 	&	0.8673 	&	0.8293 	&	0.8853 	&	0.8263 	&	0.8721 	&	0.8387 	&	0.8288 	&	0.5692 	&	0.5538 	&	0.9233 	&	0.9340 	&	0.5413 	&	0.6057 	&	0.8892 	&	0.9395 	&	0.9136 	&	0.9329 	&	0.8512 	&	0.8407 	&	0.0768 	&	0.0721 	\\
$\spadesuit$ Dover \cite{wu2023exploring}	&	0.7854 	&	0.8206 	&	0.7071 	&	0.7781 	&	0.8792 	&	0.9035 	&	0.7451 	&	0.7208 	&	0.2846 	&	0.3022 	&	0.9256 	&	0.9363 	&	0.5712 	&	0.5882 	&	0.8848 	&	0.9155 	&	0.8605 	&	0.9023 	&	0.8600 	&	0.8323 	&	0.0179 	&	0.0523 	\\
$\spadesuit$ KSVQE \cite{lu2024kvq}&	0.8033 	&	0.8681 	&	0.7853 	&	0.8789 	&	0.8538 	&	0.9286 	&	0.8275 	&	0.8111 	&	0.4415 	&	0.4324 	&	0.9320 	&	0.9545 	&	0.6533 	&	0.7616 	&	0.9032 	&	0.9572 	&	0.8948 	&	0.9533 	&	0.8704 	&	0.8664 	&	0.2063 	&	0.2333 	\\

\hdashline
\rowcolor[gray]{.92}
\textbf{DynaMesh-Rater}	&	0.8169	& 0.8951 & 0.8032 & 0.9080 & 0.8819 & 0.9483 & 0.8517 & 0.8241 & 0.5463 & 0.5494 & 0.9159 & 0.9561 & 0.6455 & 0.7715 & 0.9270 & 0.9711 & 0.9154	& 0.9594 & 0.8894	& 0.8845 & 0.4349	& 0.3884 \\

\bottomrule
\end{tabular}
}
\label{tab:exp_textsubset_mos}
\end{table*}

\subsection{Comparison Methods}

We select various types of quality assessment methods for benchmarking our DHQA-4D dataset, including (1) full reference video quality assessment methods: PSNR$_{rgb}$, PSNR$_{yuv}$, SSIM, MS-SSIM, GMSD and G-LPIPS \cite{nehme2023textured}; (2) no-reference image quality assessment methods: HyperNet \cite{su2020blindly}, MUSIQ \cite{ke2021musiq} and MANIQA \cite{yang2022maniqa}; (3) no-reference video quality assessment methods: VSFA, GSTVQA, SimpleVQA, FastVQA, Dover and KSVQE \cite{lu2024kvq}. For performance evaluation, we select Spearman
rank-order correlation coefficient (SRCC), Kendall rank-order correlation coefficient (KRCC), and Pearson linear
correlation coefficient (PLCC) to measure the
performance of all quality assessment models.

\begin{table}[h]
\centering
\renewcommand\arraystretch{1.7}
\caption{ \textbf{Benchmark performance} on the non-textured mesh subset of the DHQA-4D dataset for different distortion types. $\heartsuit$, $\clubsuit$, and $\spadesuit$ denote full reference quality assessment method, no-reference image quality assessment method, and no-reference video quality assessment method, respectively.}
\resizebox{\linewidth}{!}{
\begin{tabular}{l|*{5}{cc}}
\toprule
\multirow{2}{*}{\makecell{\textbf{Method}}} 
& \multicolumn{2}{c}{\textbf{GN}} & \multicolumn{2}{c}{\textbf{MS}} 
& \multicolumn{2}{c}{\textbf{PC}} & \multicolumn{2}{c}{\textbf{MC}} 
& \multicolumn{2}{c}{\textbf{DC}} \\
\cmidrule(lr){2-3} \cmidrule(lr){4-5} \cmidrule(lr){6-7} \cmidrule(lr){8-9} \cmidrule(lr){10-11}
& SRCC$\uparrow$ & PLCC$\uparrow$
& SRCC$\uparrow$ & PLCC$\uparrow$
& SRCC$\uparrow$ & PLCC$\uparrow$
& SRCC$\uparrow$ & PLCC$\uparrow$
& SRCC$\uparrow$ & PLCC$\uparrow$ \\
\midrule

$\heartsuit$ PSNR$_{rgb}$	&	0.8753 	&	0.9227 	&	0.9176 	&	0.8761 	&	0.8689 	&	0.8820 	&	0.9075 	&	0.9168 	&	0.3891 	&	0.2006 	\\
$\heartsuit$ PSNR$_{yuv}$ 	&	0.8757 	&	0.9408 	&	0.9072 	&	0.9213 	&	0.8689 	&	0.9020 	&	0.9072 	&	0.9213 	&	0.3891 	&	0.5197 	\\
$\heartsuit$ SSIM \cite{wang2004image}	&	0.8820 	&	0.7700 	& 0.9193 	&	0.7915 	&	0.9033 	&	0.8824 	&	0.9170 	&	0.8917 	&	0.3520 	&	0.2085 	\\
$\heartsuit$ MS-SSIM \cite{wang2003multiscale} 	&	0.8823 	&	0.7569 	&	0.9181 	&	0.7949 	&	0.9028 	&	0.8897 	&	0.9153 	&	0.8879 	&	0.3035 	&	0.1967 	\\
$\heartsuit$ GMSD \cite{xue2013gradient}	&	0.8844 	&	0.8854 	&	0.9213 	&	0.8656 	&	0.8975 	&	0.9120 	&	0.9139 	&	0.9213 	&	0.4200 	&	0.2195 	\\
$\heartsuit$ G-LPIPS \cite{nehme2023textured} 	&	0.8858 	&	0.9273 	&	0.9187 	&	0.8996 	&	0.9141 	&	0.9597 	&	0.9179 	&	0.9208 	&	0.2796 	&	0.3790 	\\

\hdashline

$\clubsuit$ HyperNet \cite{su2020blindly} &  0.8125 & 0.7396  & 0.7646  & 0.7062  & 0.8607  & 0.8379  & 0.8901  & 0.8807  & 0.0601  & 0.1626  \\
$\clubsuit$ MUSIQ \cite{ke2021musiq} & 0.8457  & 0.8516 & 0.8490  & 0.8326  & 0.8711  & 0.8749  & 0.8985  & 0.8964  & 0.0431  & 0.0625  \\
$\clubsuit$ MANIQA \cite{yang2022maniqa} & 0.8636 & 0.8840 & 0.8741 & 0.8573 &  0.9016
&  0.9398 & 0.8987 & 0.9023 &  0.0802 &  0.0471 \\

\hdashline

$\spadesuit$ VSFA \cite{li2019quality} & 0.8847  & 0.9014  & 0.8792  & 0.8733  & 0.8826  & 0.9274  & 0.8765  & 0.8874  &  0.1342 & 0.1754  \\
$\spadesuit$ GSTVQA \cite{chen2021learning} &  0.9069 & 0.9085  & 0.8742  & 0.8706  & 0.8967  & 0.9380  & 0.8945  & 0.9024  & 0.1748  & 0.1653  \\
$\spadesuit$ SimpleVQA \cite{sun2022deep} & 0.9015  & 0.9126  & 0.8985  & 0.8959  & 0.9063  & 0.9508  & 0.9011  & 0.9118  & 0.2278  & 0.2275  \\

$\spadesuit$ FastVQA \cite{wu2022fast}	&	0.9137 	&	0.8985 	&	0.8878 	&	0.8506 	&	0.9037 	&	0.9302 	&	0.8880 	&	0.8770 	&	0.1737 	&	0.1670 	\\
$\spadesuit$ Dover \cite{wu2023exploring}	&	0.9040 	&	0.9059 	&	0.8708 	&	0.8263 	&	0.9095 	&	0.9515 	&	0.9026 	&	0.8952 	&	0.1872 	&	0.1628 	\\
$\spadesuit$ KSVQE \cite{lu2024kvq}	&	0.9093	&	0.9117 	&	0.9180 	&	0.8992 	&	0.9042 	&	0.9483 	&	0.9214 	&	0.9223 	&	0.1723 	&	0.1747 	\\

\hdashline
\rowcolor[gray]{.92}
\textbf{DynaMesh-Rater} & 0.8991 &	0.9414 & 0.8979	& 0.8978 & 0.9104 & 0.9547 & 0.9117 &	0.9103 & 0.4616	& 0.4302 \\

\bottomrule
\end{tabular}
}
\vspace{-1mm}
\label{tab:exp_shapesubset_mos}
\end{table}

\subsection{Implementation Details} Our framework is based on the PyTorch framework. The batch size for both subsets is set to 8, and all the models are trained for 8 epochs. To extract video features, we select InternVIT \cite{chen2024internvl} as the visual encoder, and the projected video is obtained using a dynamic camera that rotates 360 degrees horizontally around the dynamic human mesh; then we uniformly sample 5 frames from the video. To extract motion features, we uniformly split the video into 5 small clips and utilize the SlowFast model \cite{feichtenhofer2019slowfast} to obtain the corresponding feature tokens for each video clip. To extract geometry features, we uniformly sample 5 static human meshes and compute the geometry features for each static mesh, then all the geometry features are fed into the LLM model. The LoRA rank parameters for both the visual encoder and LLM model are set to 12.

\subsection{Performance and Analysis}

We summarized the quality assessment results on both the textured mesh subset and the non-textured mesh subset in Table \ref{tab:exp_overall_mos}, Table \ref{tab:exp_textsubset_mos}, and Table \ref{tab:exp_shapesubset_mos}. From the Table \ref{tab:exp_overall_mos}, it can be observed that our DynaMesh-Rater achieves the best performance in terms of SRCC, PLCC, and KRCC on both the textured mesh subset and the non-textured mesh subset, demonstrating the effectiveness of the LMM model for the quality assessment task. Meanwhile, we can observe that no reference video quality assessment methods generally achieve better performance compared to full reference methods and no reference image quality assessment methods. The reason is that video quality assessment methods can naturally capture the dynamic characteristics of 4D meshes. Among all image quality assessment models, MANIQA \cite{yang2022maniqa} achieves the best performance. Among all video quality assessment methods, KSVQE \cite{lu2024kvq} achieves the best performance on both subsets.

In addition to the overall score analysis, Table \ref{tab:exp_textsubset_mos} and Table \ref{tab:exp_shapesubset_mos} report the distortion-specific performance on both subsets. We can observe that current methods are difficult to predict accurate quality in terms of TMC, UMC, and DC distortion types. For TMC and UMC distortions, the reason might be that these two types of noise do not have such obvious visual distortion for the mesh, so the current methods find it difficult to distinguish them. For DC distortion, the reason lies in that the current methods are unable to precisely perceive the temporal distortion of the dynamic mesh.

\begin{table}
\renewcommand\arraystretch{1.7}
  \caption{Ablation study of different components in our DynaMesh-Rater method. ``Visual", ``Motion" and ``Geometry" denote visual feature, motion feature, and geometry feature, respectively.}
  \resizebox{\linewidth}{!}{
  \begin{tabular}{cccccc| ccc ccc}
    \toprule
                                   & \multicolumn{5}{c}{Feature \& Strategy}                                  & \multicolumn{3}{c}{Textured Mesh}                   & \multicolumn{3}{c}{Non-Textured Mesh} \\
    \cmidrule(r){2-6} \cmidrule(r){7-9} \cmidrule(r){10-12} 
\multicolumn{1}{c}{No.}   & Visual  & Motion & Geometry & LoRA$_{vis}$ & LoRA$_{llm}$ & SRCC     & PLCC     & KRCC          & SRCC       & PLCC    & KRCC        \\

\multicolumn{1}{c}{(1)}   &  \ding{52}  &   &    &    &    &  0.8520 & 0.8543 & 0.7069 & 0.8423 & 0.8536 & 0.7168 \\

\multicolumn{1}{c}{(2)}  &  \ding{52}  &   &   & \ding{52}  &    &  0.8823 & 0.8875 & 0.7319 & 0.8846 & 0.8852 & 0.7417 \\

\multicolumn{1}{c}{(3)}   &  \ding{52}  &   &    & \ding{52} &  \ding{52}  & 0.9087 & 0.9104 & 0.7553 & 0.9165  & 0.9106  & 0.7641  \\

\multicolumn{1}{c}{(4)}   &  \ding{52}  &  \ding{52}  &    & \ding{52} &  \ding{52} & 0.9269 & 0.9244 & 0.7659 &  0.9201 & \textbf{0.9195} & 0.7652  \\

\rowcolor[gray]{.92}
\multicolumn{1}{c}{(5)}   &  \ding{52}  &  \ding{52}  & \ding{52}   & \ding{52} &  \ding{52} & \textbf{0.9316} & \textbf{0.9327} & \textbf{0.7762} &  \textbf{0.9217} & 0.9161 & \textbf{0.7713} \\
 
    \bottomrule
  \end{tabular}\label{ablation}
   }
  \centering
  \label{ablation}
\end{table}

\subsection{Ablation Study}

In this section, we conduct an ablation study to validate the effectiveness of the different components of our DynaMesh-Rater method. We report the experiment results on both the textured mesh subset and non-textured mesh subset of our DHQA-4D dataset as demonstrated in Table \ref{ablation}. 

\subsubsection{Effectiveness of LoRA Adaptation.} First, we divide the LoRA adaptation into LoRA for the vision encoder and LoRA for the language model. Then, we study the performance impact of these two LoRA on predicting the quality score of dynamic meshes. The experimental results show that both LoRA for the vision encoder and LoRA for the language model can effectively improve the score performance of quality evaluation.

\subsubsection{Effectiveness of Multi-dimensional Features.} Table \ref{ablation} clearly shows that the combination of all multi-modal features achieves the best performance result. The last three rows demonstrate the importance of motion features and geometry features, which sequentially improve quality assessment performance. The motion features can help the neural network to precisely perceive the change of movement, especially the temporal discontinuity type. The geometry features can provide additional complementary information beyond visual information, which is very important for the quality evaluation of dynamic meshes.

\section{Conclusion}

In this paper, we first propose a large-scale dynamic \underline{d}igital \underline{h}uman \underline{q}uality \underline{a}ssessment dataset, DHQA-4D which contains 32 high-quality real-scanned 4D human mesh sequences, 1920 distorted textured 4D human meshes degraded by 11 textured distortions, as well as their corresponding textured and non-textured mean opinion scores (MOSs). Equipped with the DHQA-4D dataset, we analyze the influence of different types of distortion on human perception for textured dynamic 4D meshes and non-textured dynamic 4D meshes. Additionally, we propose DynaMesh-Rater, a novel large multimodal model (LMM) based approach that is able to assess both textured 4D meshes and non-textured 4D meshes. Concretely, DynaMesh-Rater elaborately extracts multi-dimensional features, including visual features from a projected 2D video, motion features from cropped video clips, and geometry features from the 4D human mesh to provide comprehensive quality-related information. Then we utilize a LMM model to integrate the multi-dimensional features and conduct a LoRA-based instruction tuning technique to teach the LMM model to predict the quality scores. Extensive experimental results on the DHQA-4D dataset demonstrate the superiority of our DynaMesh-Rater method over previous quality assessment methods.

\bibliographystyle{IEEEtran}
\bibliography{refs}

\begin{thebibliography}{10}
\providecommand{\url}[1]{#1}
\csname url@samestyle\endcsname
\providecommand{\newblock}{\relax}
\providecommand{\bibinfo}[2]{#2}
\providecommand{\BIBentrySTDinterwordspacing}{\spaceskip=0pt\relax}
\providecommand{\BIBentryALTinterwordstretchfactor}{4}
\providecommand{\BIBentryALTinterwordspacing}{\spaceskip=\fontdimen2\font plus
\BIBentryALTinterwordstretchfactor\fontdimen3\font minus \fontdimen4\font\relax}
\providecommand{\BIBforeignlanguage}[2]{{%
\expandafter\ifx\csname l@#1\endcsname\relax
\typeout{** WARNING: IEEEtran.bst: No hyphenation pattern has been}%
\typeout{** loaded for the language `#1'. Using the pattern for}%
\typeout{** the default language instead.}%
\else
\language=\csname l@#1\endcsname
\fi
#2}}
\providecommand{\BIBdecl}{\relax}
\BIBdecl

\bibitem{nehme2020visual}
Y.~Nehm{\'e}, F.~Dupont, J.-P. Farrugia, P.~Le~Callet, and G.~Lavou{\'e}, ``Visual quality of 3d meshes with diffuse colors in virtual reality: Subjective and objective evaluation,'' \emph{IEEE Transactions on Visualization and Computer Graphics}, vol.~27, no.~3, pp. 2202--2219, 2020.

\bibitem{nehme2023textured}
Y.~Nehm{\'e}, J.~Delanoy, F.~Dupont, J.-P. Farrugia, P.~Le~Callet, and G.~Lavou{\'e}, ``Textured mesh quality assessment: Large-scale dataset and deep learning-based quality metric,'' \emph{ACM Transactions on Graphics}, vol.~42, no.~3, pp. 1--20, 2023.

\bibitem{yang2024tdmd}
Q.~Yang, J.~Jung, T.~Deschamps, X.~Xu, and S.~Liu, ``Tdmd: A database for dynamic color mesh quality assessment study,'' \emph{IEEE Transactions on Visualization and Computer Graphics}, 2024.

\bibitem{zhang2023ddh}
Z.~Zhang, Y.~Zhou, W.~Sun, W.~Lu, X.~Min, Y.~Wang, and G.~Zhai, ``Ddh-qa: A dynamic digital humans quality assessment database,'' in \emph{2023 IEEE International Conference on Multimedia and Expo (ICME)}.\hskip 1em plus 0.5em minus 0.4em\relax IEEE, 2023, pp. 2519--2524.

\bibitem{torkhani2015perceptual}
F.~Torkhani, K.~Wang, and J.-M. Chassery, ``Perceptual quality assessment of 3d dynamic meshes: Subjective and objective studies,'' \emph{Signal Processing: Image Communication}, vol.~31, pp. 185--204, 2015.

\bibitem{cui2024sjtu}
B.~Cui, Q.~Yang, K.~Yang, Y.~Xu, X.~Xu, and S.~Liu, ``Sjtu-tmqa: A quality assessment database for static mesh with texture map,'' in \emph{ICASSP 2024-2024 IEEE international conference on acoustics, speech and signal processing (ICASSP)}.\hskip 1em plus 0.5em minus 0.4em\relax IEEE, 2024, pp. 7875--7879.

\bibitem{zhang2023advancing}
Z.~Zhang, W.~Sun, Y.~Zhou, H.~Wu, C.~Li, X.~Min, X.~Liu, G.~Zhai, and W.~Lin, ``Advancing zero-shot digital human quality assessment through text-prompted evaluation,'' \emph{arXiv preprint arXiv:2307.02808}, 2023.

\bibitem{lavoue2006perceptually}
G.~Lavou{\'e}, E.~Drelie~Gelasca, F.~Dupont, A.~Baskurt, and T.~Ebrahimi, ``Perceptually driven 3d distance metrics with application to watermarking,'' \emph{Applications of Digital Image Processing XXIX}, vol. 6312, p. 63120L, 2006.

\bibitem{corsini2007watermarked}
M.~Corsini, E.~D. Gelasca, T.~Ebrahimi, and M.~Barni, ``Watermarked 3-d mesh quality assessment,'' \emph{IEEE Transactions on Multimedia}, vol.~9, no.~2, pp. 247--256, 2007.

\bibitem{christaki2019subjective}
K.~Christaki, E.~Christakis, P.~Drakoulis, A.~Doumanoglou, N.~Zioulis, D.~Zarpalas, and P.~Daras, ``Subjective visual quality assessment of immersive 3d media compressed by open-source static 3d mesh codecs,'' in \emph{MultiMedia Modeling: 25th International Conference, MMM 2019, Thessaloniki, Greece, January 8--11, 2019, Proceedings, Part I 25}.\hskip 1em plus 0.5em minus 0.4em\relax Springer, 2019, pp. 80--91.

\bibitem{guo2016subjective}
J.~Guo, V.~Vidal, I.~Cheng, A.~Basu, A.~Baskurt, and G.~Lavoue, ``Subjective and objective visual quality assessment of textured 3d meshes,'' \emph{ACM Transactions on Applied Perception (TAP)}, vol.~14, no.~2, pp. 1--20, 2016.

\bibitem{gao2025ges}
Z.~Gao, Y.~Li, S.~Wu, Y.~Cao, H.~Duan, and G.~Zhai, ``Ges-qa: A multidimensional quality assessment dataset for audio-to-3d gesture generation,'' \emph{arXiv preprint arXiv:2508.12020}, 2025.

\bibitem{mekuria2016evaluation}
R.~Mekuria, Z.~Li, C.~Tulvan, and P.~Chou, ``Evaluation criteria for pcc (point cloud compression),'' \emph{ISO/IEC JTC}, vol.~1, p. N16332, 2016.

\bibitem{tian2017geometric}
D.~Tian, H.~Ochimizu, C.~Feng, R.~Cohen, and A.~Vetro, ``Geometric distortion metrics for point cloud compression,'' in \emph{2017 IEEE International Conference on Image Processing (ICIP)}, 2017, pp. 3460--3464.

\bibitem{alexiou2018point}
E.~Alexiou and T.~Ebrahimi, ``Point cloud quality assessment metric based on angular similarity,'' in \emph{IEEE International Conference on Multimedia and Expo}, 2018, pp. 1--6.

\bibitem{torlig2018novel}
E.~M. Torlig, E.~Alexiou, T.~A. Fonseca, R.~L. de~Queiroz, and T.~Ebrahimi, ``A novel methodology for quality assessment of voxelized point clouds,'' in \emph{Applications of Digital Image Processing XLI}, vol. 10752, 2018, pp. 174--190.

\bibitem{yang2020graphsim}
Q.~Yang, Z.~Ma, Y.~Xu, Z.~Li, and J.~Sun, ``Inferring point cloud quality via graph similarity,'' \emph{IEEE Transactions on Pattern Analysis and Machine Intelligence}, 2020.

\bibitem{meynet2020pcqm}
G.~Meynet, Y.~Nehm{\'e}, J.~Digne, and G.~Lavou{\'e}, ``Pcqm: A full-reference quality metric for colored 3d point clouds,'' in \emph{2020 Twelfth International Conference on Quality of Multimedia Experience (QoMEX)}, 2020, pp. 1--6.

\bibitem{alexiou2020pointssim}
E.~Alexiou and T.~Ebrahimi, ``Towards a point cloud structural similarity metric,'' in \emph{2020 IEEE International Conference on Multimedia \& Expo Workshops (ICMEW)}, 2020, pp. 1--6.

\bibitem{pcqa-large-scale}
Y.~Liu, Q.~Yang, Y.~Xu, and L.~Yang, ``Point cloud quality assessment: Dataset construction and learning-based no-reference metric,'' \emph{ACM Transactions on Multimedia Computing, Communications, and Applications (TOMM)}, 2022.

\bibitem{zhang2022no}
Z.~Zhang, W.~Sun, X.~Min, T.~Wang, W.~Lu, and G.~Zhai, ``No-reference quality assessment for 3d colored point cloud and mesh models,'' \emph{IEEE Transactions on Circuits and Systems for Video Technology}, 2022.

\bibitem{zhou2022blind}
W.~Zhou, Q.~Yang, Q.~Jiang, G.~Zhai, and W.~Lin, ``Blind quality assessment of 3d dense point clouds with structure guided resampling,'' \emph{arXiv preprint arXiv:2208.14603}, 2022.

\bibitem{wang2024zoom}
J.~Wang, W.~Gao, and G.~Li, ``Zoom to perceive better: No-reference point cloud quality assessment via exploring effective multiscale feature,'' \emph{IEEE Transactions on Circuits and Systems for Video Technology}, vol.~34, no.~7, pp. 6334--6346, 2024.

\bibitem{yang2020predicting}
Q.~Yang, H.~Chen, Z.~Ma, Y.~Xu, R.~Tang, and J.~Sun, ``Predicting the perceptual quality of point cloud: A 3d-to-2d projection-based exploration,'' \emph{IEEE transactions on multimedia}, vol.~23, pp. 3877--3891, 2020.

\bibitem{chai2024plain}
X.~Chai, F.~Shao, B.~Mu, H.~Chen, Q.~Jiang, and Y.-S. Ho, ``Plain-pcqa: No-reference point cloud quality assessment by analysis of plain visual and geometrical components,'' \emph{IEEE Transactions on Circuits and Systems for Video Technology}, vol.~34, no.~7, pp. 6207--6223, 2024.

\bibitem{zhang2021mesh}
Z.~Zhang, W.~Sun, X.~Min, T.~Wang, W.~Lu, W.~Zhu, and G.~Zhai, ``A no-reference visual quality metric for 3d color meshes,'' in \emph{2021 IEEE International Conference on Multimedia \& Expo Workshops (ICMEW)}.\hskip 1em plus 0.5em minus 0.4em\relax IEEE, 2021, pp. 1--6.

\bibitem{fan2022no}
Y.~Fan, Z.~Zhang, W.~Sun, X.~Min, N.~Liu, Q.~Zhou, J.~He, Q.~Wang, and G.~Zhai, ``A no-reference quality assessment metric for point cloud based on captured video sequences,'' in \emph{2022 IEEE 24th International Workshop on Multimedia Signal Processing (MMSP)}.\hskip 1em plus 0.5em minus 0.4em\relax IEEE, 2022, pp. 1--5.

\bibitem{zhang2022treating}
Z.~Zhang, W.~Sun, X.~Min, W.~Wu, Y.~Chen, and G.~Zhai, ``Treating point cloud as moving camera videos: A no-reference quality assessment metric,'' \emph{arXiv preprint arXiv:2208.14085}, 2022.

\bibitem{liu2021pqa}
Q.~Liu, H.~Yuan, H.~Su, H.~Liu, Y.~Wang, H.~Yang, and J.~Hou, ``Pqa-net: Deep no reference point cloud quality assessment via multi-view projection,'' \emph{IEEE Transactions on Circuits and Systems for Video Technology}, vol.~31, no.~12, pp. 4645--4660, 2021.

\bibitem{chen2024dynamic}
W.~Chen, Q.~Jiang, W.~Zhou, L.~Xu, and W.~Lin, ``Dynamic hypergraph convolutional network for no-reference point cloud quality assessment,'' \emph{IEEE Transactions on Circuits and Systems for Video Technology}, 2024.

\bibitem{ge2024lmm}
Q.~Ge, W.~Sun, Y.~Zhang, Y.~Li, Z.~Ji, F.~Sun, S.~Jui, X.~Min, and G.~Zhai, ``Lmm-vqa: Advancing video quality assessment with large multimodal models,'' \emph{arXiv preprint arXiv:2408.14008}, 2024.

\bibitem{wu2023q}
H.~Wu, Z.~Zhang, W.~Zhang, C.~Chen, L.~Liao, C.~Li, Y.~Gao, A.~Wang, E.~Zhang, W.~Sun \emph{et~al.}, ``Q-align: Teaching lmms for visual scoring via discrete text-defined levels,'' \emph{arXiv preprint arXiv:2312.17090}, 2023.

\bibitem{zhang2024human}
Z.~Zhang, W.~Sun, X.~Li, Y.~Li, Q.~Ge, J.~Jia, Z.~Zhang, Z.~Ji, F.~Sun, S.~Jui \emph{et~al.}, ``Human-activity agv quality assessment: A benchmark dataset and an objective evaluation metric,'' \emph{arXiv preprint arXiv:2411.16619}, 2024.

\bibitem{wu2023exploring}
H.~Wu, E.~Zhang, L.~Liao, C.~Chen, J.~Hou, A.~Wang, W.~Sun, Q.~Yan, and W.~Lin, ``Exploring video quality assessment on user generated contents from aesthetic and technical perspectives,'' in \emph{Proceedings of the IEEE/CVF International Conference on Computer Vision}, 2023, pp. 20\,144--20\,154.

\bibitem{li2025aghi}
Y.~Li, S.~Wu, W.~Sun, Z.~Zhang, Y.~Zhu, Z.~Zhang, H.~Duan, X.~Min, and G.~Zhai, ``Aghi-qa: A subjective-aligned dataset and metric for ai-generated human images,'' \emph{arXiv preprint arXiv:2504.21308}, 2025.

\bibitem{wu2025fvq}
S.~Wu, Y.~Li, Z.~Xu, Y.~Gao, H.~Duan, W.~Sun, and G.~Zhai, ``Fvq: A large-scale dataset and a lmm-based method for face video quality assessment,'' \emph{arXiv preprint arXiv:2504.09255}, 2025.

\bibitem{zhou2026mi3s}
Y.~Zhou, Z.~Zhang, S.~Wu, J.~Jia, Y.~Jiang, W.~Sun, X.~Liu, X.~Min, and G.~Zhai, ``Mi3s: A multimodal large language model assisted quality assessment framework for ai-generated talking heads,'' \emph{Information Processing \& Management}, vol.~63, no.~1, p. 104321, 2026.

\bibitem{zhang2024q}
Z.~Zhang, H.~Wu, E.~Zhang, G.~Zhai, and W.~Lin, ``Q-bench: A benchmark for multi-modal foundation models on low-level vision from single images to pairs,'' \emph{IEEE Transactions on Pattern Analysis and Machine Intelligence}, 2024.

\bibitem{duan2024finevq}
H.~Duan, Q.~Hu, J.~Wang, L.~Yang, Z.~Xu, L.~Liu, X.~Min, C.~Cai, T.~Ye, X.~Zhang \emph{et~al.}, ``Finevq: Fine-grained user generated content video quality assessment,'' \emph{arXiv preprint arXiv:2412.19238}, 2024.

\bibitem{zhang2024lmm}
Z.~Zhang, H.~Wu, Y.~Zhou, C.~Li, W.~Sun, C.~Chen, X.~Min, X.~Liu, W.~Lin, and G.~Zhai, ``Lmm-pcqa: Assisting point cloud quality assessment with lmm,'' in \emph{Proceedings of the 32nd ACM International Conference on Multimedia}, 2024, pp. 7783--7792.

\bibitem{wang20244d}
W.~Wang, H.-I. Ho, C.~Guo, B.~Rong, A.~Grigorev, J.~Song, J.~J. Zarate, and O.~Hilliges, ``4d-dress: A 4d dataset of real-world human clothing with semantic annotations,'' in \emph{Proceedings of the IEEE/CVF Conference on Computer Vision and Pattern Recognition}, 2024, pp. 550--560.

\bibitem{liu2022perceptual}
Q.~Liu, H.~Su, Z.~Duanmu, W.~Liu, and Z.~Wang, ``Perceptual quality assessment of colored 3d point clouds,'' \emph{IEEE Transactions on Visualization and Computer Graphics}, vol.~29, no.~8, pp. 3642--3655, 2022.

\bibitem{fairchild2013color}
M.~D. Fairchild, \emph{Color appearance models}.\hskip 1em plus 0.5em minus 0.4em\relax John Wiley \& Sons, 2013.

\bibitem{hasler2003measuring}
D.~Hasler and S.~E. Suesstrunk, ``Measuring colorfulness in natural images,'' in \emph{Human vision and electronic imaging VIII}, vol. 5007.\hskip 1em plus 0.5em minus 0.4em\relax SPIE, 2003, pp. 87--95.

\bibitem{zhou2024subjective}
Y.~Zhou, Z.~Zhang, W.~Sun, X.~Liu, X.~Min, and G.~Zhai, ``Subjective and objective quality-of-experience assessment for 3d talking heads,'' in \emph{Proceedings of the 32nd ACM International Conference on Multimedia}, 2024, pp. 6033--6042.

\bibitem{series2012methodology}
B.~Series, ``Methodology for the subjective assessment of the quality of television pictures,'' \emph{Recommendation ITU-R BT}, vol. 500, no.~13, 2012.

\bibitem{chen2024internvl}
Z.~Chen, J.~Wu, W.~Wang, W.~Su, G.~Chen, S.~Xing, M.~Zhong, Q.~Zhang, X.~Zhu, L.~Lu \emph{et~al.}, ``Internvl: Scaling up vision foundation models and aligning for generic visual-linguistic tasks,'' in \emph{Proceedings of the IEEE/CVF conference on computer vision and pattern recognition}, 2024, pp. 24\,185--24\,198.

\bibitem{feichtenhofer2019slowfast}
C.~Feichtenhofer, H.~Fan, J.~Malik, and K.~He, ``Slowfast networks for video recognition,'' in \emph{Proceedings of the IEEE/CVF international conference on computer vision}, 2019, pp. 6202--6211.

\bibitem{wang2004image}
Z.~Wang, A.~C. Bovik, H.~R. Sheikh, and E.~P. Simoncelli, ``Image quality assessment: from error visibility to structural similarity,'' \emph{IEEE transactions on image processing}, vol.~13, no.~4, pp. 600--612, 2004.

\bibitem{wang2003multiscale}
Z.~Wang, E.~P. Simoncelli, and A.~C. Bovik, ``Multiscale structural similarity for image quality assessment,'' in \emph{The Thrity-Seventh Asilomar Conference on Signals, Systems \& Computers, 2003}, vol.~2.\hskip 1em plus 0.5em minus 0.4em\relax Ieee, 2003, pp. 1398--1402.

\bibitem{xue2013gradient}
W.~Xue, L.~Zhang, X.~Mou, and A.~C. Bovik, ``Gradient magnitude similarity deviation: A highly efficient perceptual image quality index,'' \emph{IEEE transactions on image processing}, vol.~23, no.~2, pp. 684--695, 2013.

\bibitem{su2020blindly}
S.~Su, Q.~Yan, Y.~Zhu, C.~Zhang, X.~Ge, J.~Sun, and Y.~Zhang, ``Blindly assess image quality in the wild guided by a self-adaptive hyper network,'' in \emph{Proceedings of the IEEE/CVF conference on computer vision and pattern recognition}, 2020, pp. 3667--3676.

\bibitem{ke2021musiq}
J.~Ke, Q.~Wang, Y.~Wang, P.~Milanfar, and F.~Yang, ``Musiq: Multi-scale image quality transformer,'' in \emph{Proceedings of the IEEE/CVF international conference on computer vision}, 2021, pp. 5148--5157.

\bibitem{yang2022maniqa}
S.~Yang, T.~Wu, S.~Shi, S.~Lao, Y.~Gong, M.~Cao, J.~Wang, and Y.~Yang, ``Maniqa: Multi-dimension attention network for no-reference image quality assessment,'' in \emph{Proceedings of the IEEE/CVF conference on computer vision and pattern recognition}, 2022, pp. 1191--1200.

\bibitem{li2019quality}
D.~Li, T.~Jiang, and M.~Jiang, ``Quality assessment of in-the-wild videos,'' in \emph{Proceedings of the 27th ACM international conference on multimedia}, 2019, pp. 2351--2359.

\bibitem{chen2021learning}
B.~Chen, L.~Zhu, G.~Li, F.~Lu, H.~Fan, and S.~Wang, ``Learning generalized spatial-temporal deep feature representation for no-reference video quality assessment,'' \emph{IEEE Transactions on Circuits and Systems for Video Technology}, vol.~32, no.~4, pp. 1903--1916, 2021.

\bibitem{sun2022deep}
W.~Sun, X.~Min, W.~Lu, and G.~Zhai, ``A deep learning based no-reference quality assessment model for ugc videos,'' in \emph{Proceedings of the 30th ACM International Conference on Multimedia}, 2022, pp. 856--865.

\bibitem{wu2022fast}
H.~Wu, C.~Chen, J.~Hou, L.~Liao, A.~Wang, W.~Sun, Q.~Yan, and W.~Lin, ``Fast-vqa: Efficient end-to-end video quality assessment with fragment sampling,'' in \emph{European conference on computer vision}.\hskip 1em plus 0.5em minus 0.4em\relax Springer, 2022, pp. 538--554.

\bibitem{lu2024kvq}
Y.~Lu, X.~Li, Y.~Pei, K.~Yuan, Q.~Xie, Y.~Qu, M.~Sun, C.~Zhou, and Z.~Chen, ``Kvq: Kwai video quality assessment for short-form videos,'' in \emph{Proceedings of the IEEE/CVF Conference on Computer Vision and Pattern Recognition}, 2024, pp. 25\,963--25\,973.

\end{thebibliography}


\vfill

\end{document}